\documentclass[10pt,journal,compsoc, twoside]{IEEEtran}
\usepackage{graphicx}
\usepackage{xcolor}
\usepackage{amstext}
\usepackage{amsmath}
\usepackage{amssymb}
\usepackage{booktabs}
\usepackage{amsthm}
\newtheorem{theorem}{Theorem}

\usepackage[flushleft]{threeparttable}
\usepackage{array, multirow, multicol}
\DeclareMathOperator{\tr}{tr}

\newcommand\norm[1]{\left\lVert#1\right\rVert}

\usepackage{tabularx,ragged2e}
\usepackage{algorithm}
\usepackage{algpseudocode}

\algnewcommand\algorithmicinput{\textbf{Input:}}
\algnewcommand\Input{\item[\algorithmicinput]}
\algnewcommand\algorithmicoutput{\textbf{Output:}}
\algnewcommand\Output{\item[\algorithmicoutput]}
\ifCLASSOPTIONcompsoc
  \usepackage[nocompress]{cite}
\else
  \usepackage{cite}
\fi
\begin{document}
%
\title{Matrix Normal PCA for Interpretable Dimension Reduction and Graphical Noise Modeling}

\author{Chihao~Zhang, ~Kuo~Gai
    and~Shihua~Zhang
    \IEEEcompsocitemizethanks{\IEEEcompsocthanksitem Chihao Zhang, Kuo Gai and Shihua Zhang* are with the NCMIS, CEMS, RCSDS, Academy of Mathematics and Systems Science, Chinese Academy of Sciences, Beijing 100190, China, and School of Mathematical Sciences, University of Chinese Academy of Sciences, Beijing 100049, China. \protect\\ *To whom correspondence should be addressed. Email: zsh@amss.ac.cn.}}

\markboth{ZHANG C, Gai K, ZHANG S.:  MATRIX NORMAL PCA FOR INTERPRETABLE DIMENSION REDUCTION AND GRAPHICAL NOISE MODELING}%
{MANUSCRIPT, VOL. XX, NO. XX, XXX 2019}
%



\IEEEtitleabstractindextext{%
\begin{abstract}
Principal component analysis (PCA) is one of the most widely used dimension reduction and multivariate statistical techniques. From a probabilistic perspective, PCA seeks a low-dimensional representation of data in the presence of independent identical Gaussian noise. Probabilistic PCA (PPCA) and its variants have been extensively studied for decades. Most of them assume the underlying noise follows a certain independent identical distribution. However, the noise in the real world is usually complicated and structured. 
To address this challenge, some variants of PCA for data with non-IID noise have been proposed. 
However, most of the existing methods only assume that the noise is correlated in the feature space while there may exist two-way structured noise.
To this end, we propose a powerful and intuitive PCA method (MN-PCA) through modeling the graphical noise by the matrix normal distribution, which enables us to explore the structure of noise in both the feature space and the sample space. MN-PCA obtains a low-rank representation of data and the structure of noise simultaneously. And it can be explained as approximating data over the generalized Mahalanobis distance. We develop two algorithms to solve this model: one maximizes the regularized likelihood, the other exploits the Wasserstein distance, which is more robust. Extensive experiments on various data demonstrate their effectiveness.
\end{abstract}

}

\maketitle

\IEEEdisplaynontitleabstractindextext

%
\IEEEpeerreviewmaketitle

\IEEEraisesectionheading{\section{Introduction}\label{sec:introduction}}

%
%
%
%
\IEEEPARstart{M}ASSIVE data emerge from diverse fields of science and engineering dramatically. For example, biologists can use microarray to detect the expression of thousands of genes at the same time; a single picture consists of hundreds, thousands, or even more pixels. Generally, those data are redundant and noisy. Directly extracting useful information from the primitive data is often infeasible. Therefore, how to discover the compact and meaningful representation of high-dimensional data becomes a fundamental problem. Researchers have developed many powerful methods to address this problem from different perspectives. Among those methods, principal component analysis (PCA) is one of the most fundamental and widely used data mining and multivariate statistical techniques. Since Pearson \cite{pearson1901liii} invented it in 1901, PCA has become a standard approach for dimension reduction and feature extraction, and has numerous applications in various fields, such as signal processing \cite{moore1981principal}, human face recognition \cite{hancock1996face} and gene expression analysis \cite{hastie2000gene}.

\par PCA obtains a low-dimensional representation for high-dimensional data in a $L_2$-norm sense. However, it is known that PCA is sensitive to gross errors. To remove the effect of sparse gross errors, robust PCA (RPCA) \cite{wright2009robust, xu2010robust, candes2011robust} has been proposed. Specifically, RPCA seeks to decompose the data matrix as a superposition of a low-rank matrix with a sparse matrix. The sparse matrix captures the gross errors, enabling RPCA to recover the low-rank representation of data accurately.
RPCA has been demonstrated to have a number of applications in various fields \cite{candes2011robust, peng2012rasl, liu2013robust, shahid2015robust}.

\par
To tackle the noise of data implicitly, regularized methods have been introduced.
A common assumption is that the representations of two data points that are close should be close too.
To preserve the local geometrical structure in the representation, graph regularizers are imposed. Gao \textit{et al.} \cite{gao2010local} proposed a sparse coding method which exploits the dependence among the feature space by constructing a Laplace matrix. Zheng \textit{et al.} \cite{zheng2011graph} also proposed a graph regularized sparse coding to learn the sparse representations that explicitly takes into account the local manifold structure of the data. More recently, Yin \textit{et al.} \cite{yin2015dual} proposed a low-rank representation method that considers the geometrical structure in both the feature space and the sample space. As many measurements in experiments are naturally nonnegative, there are also some nonnegative matrix factorization variants with graph regularization \cite{cai2011graph, zhang2011novel}.

\par Probabilistic approaches are natural ways to account for different types of noise in data.
Tipping and Bishop \cite{tipping1999probabilistic} first introduced a probabilistic PCA (PPCA) method, and showed that PCA could be derived from a Gaussian latent variable model.
PPCA assumes that the distribution of noise is independent and identical (IID) Gaussian distribution.
This probabilistic framework allows various extensions of PCA.
For example, Bishop extended PCA by a Bayesian method \cite{bishop1999bayesian} (BPCA) that can  automatically determine the number of retained principal components.
Despite the Gaussian distribution, researchers have also introduced different distributions to handle different types of noise \cite{zhao2006probabilistic, wang2012probabilistic, li2010simple}.
For example, Wang \textit{et al.} \cite{wang2012probabilistic} used the Laplace distribution to model data with gross errors.
Li and Tao \cite{li2010simple} employed the exponential family distributions to handle general types of noise.
Most of those variants assume the distribution of noise is IID.

\par However, the noise in the real-world is usually complicated and structured.
The IID assumption no longer holds.
For example, data collected from different sources may contain heterogeneous noise.
Zhang and Zhang \cite{zhang2017bayesian} developed a Bayesian joint matrix decomposition method (BJMD) to model the heterogeneous noise of multi-view data by the Gaussian distribution.
Some authors relaxed the identical assumption and introduced mixture of distributions to model the complex noise \cite{zhao2014robust, Cao_2015_ICCV}.
Beyond the independent assumption, some researchers also explored the non-IID assumption in the sense of PCA \cite{Kalaitzis2012, Han2013, Vaswani2016, Gu2020}. 
Kalaitzis and Lawrence \cite{Kalaitzis2012} generalized PPCA by adding a fixed random effect to decompose the residual variance (RCA).
RCA can be interpreted as the IID Gaussian noise with a linear offset in the feature space.
Vaswani and Guo \cite{Vaswani2016} studied the correlated-PCA problem where the noise is data-dependent. 
\par
Most of the existing methods assume that the noise is only correlated in the feature space. 
However,  the noise can be correlated among features and among samples simultaneously.
Take the gene expression data  $Y\in R^{n\times p}$ as an example, where $n$ is the number of biological samples, and $p$ is the number of genes (features).
The noise of gene expression of samples may demonstrate different patterns under different biological conditions, while the noise of genes related with the same biological processes are also correlated.
There are only a few works that model the noise structure in feature and sample spaces simultaneously.
Allen \textit{et al.} \cite{Allen2014} proposed the generalized PCA (GPCA) using the matrix normal distribution to model noise.
GPCA finds the best low-rank approximation of data with respect to the predefined sample and feature precision matrices.
The authors suggested two empirical ways to construct the precision matrices to account for the spatial-temporal relationship. 
GPCA works well for spatial-temporal data, but the predefined precision matrices can make it restrictive for general data.     
\par To this end, we propose a powerful and intuitive PCA model (MN-PCA) to obtain the low-rank representation and the structure of the underlying noise at the same time.
Different for GPCA, MN-PCA infers the precision matrices of the matrix normal distribution from the observed data.
The inference of the precision matrices of the matrix normal distribution turns to be that of two Gaussian graphical models, enabling us to explore the structure of the noise in both the feature space and the sample space.
We develop two algorithms to solve this model: one is to maximize the regularized likelihood directly; the other is to minimize the discrepancy between the distribution of the white matrix normal and that of the residue.
We extensively compare the two algorithms and discuss their advantages and disadvantages. Extensive experiments on various data show their effectiveness. 
By considering the structure of the underlying noise, MN-PCA generally obtains a better low-rank representation, and the inferred structure of the noise is interpretable and reveals some interesting information that is ignored by other methods.
\par 
The contribution of this paper is twofold. 
First, we propose an intuitive framework MN-PCA that obtains the low-rank representation and discovers the noise structure in the sample and feature spaces at the same time;
MN-PCA connects two important statistic topics, PCA and sparse precision matrix estimation.  
Second, we develop two effective algorithms from different perspectives; 
the latter algorithm based on minimizing the Wasserstein distance is more robust than that maximizes the regularized likelihood.   
This may inspire the future research of PCA for data with complicated noise.
\section{Related Work}
\subsection{PCA}
\par There are two common formulations of PCA that give rise to the same algorithm \cite{Bishop2007PatternRA}. From a dimension reduction perspective, PCA seeks a low-dimensional sub-space in which the projected variance of the data is maximized. Specifically, let $Y\in R^{n\times p}$ be the data matrix of $n$ observations and $p$ variables. For simplicity, assume the data matrix $Y$ is already centered. 
The first principal component $v_1$ is defined as
\begin{equation}
v_1 = \mathop{\arg\max}_{v} v^T S v  \quad  \text{  subject to } \norm{v} = 1,
\end{equation}
where $S= Y^TY/n$ is the $p\times p$ covariance matrix. The next principal components $v_{r+1}$ are defined in sequence:
\begin{equation}
v_{r+1}  = \mathop{\arg\max}_{v} v^T S v,
\end{equation}
subject to $\norm{v} = 1, v^Tv_l = 0, \forall 1\leq l \leq r$.
\par This definition implies that the first $r$ principal components are the first $r$ eigenvectors of $S$. Thus, we can use the singular value decomposition (SVD) to perform PCA.
Let the SVD of $Y$ be
\begin{equation}
Y = U\Sigma V^T,
\end{equation}
where $U\in R^{n\times p}$ and $V\in R^{p\times p}$ are orthogonal matrices, $\Sigma \in R^{p\times p}$ is a diagonal matrix with the diagonal elements $\sigma_i$ in descending order.
By $XV=U\Sigma$, we see that $z_r = U_{\cdot r}\sigma_r$.

\par  PCA can also be interpreted as minimizing the reconstruction error. Given $Y$, we aim at finding a low-rank approximation of $Y\approx XW^T$ under the Frobenius norm:
\begin{equation}
    \min ||Y-XW^T||_F^2 \label{eq:obj_pca},
\end{equation}
where $X\in R^{n \times k}$ and $W \in R^{p\times k}$. Eckart and Young \cite{eckart1936approximation} showed that truncated SVD has the optimality property:
\begin{equation}
||Y-Y_k||_F \leq ||Y-B||_F,   \label{eq:optimal}
\end{equation}
where $Y_k = U_k \Sigma_k V_k^T$ is the truncated SVD of $Y$, and $B$ is any matrix of rank at most $k$. Thus, $Y_k$ is the best rank-$k$ approximation of $Y$ under the Frobenius norm. Then $X=U_k \Sigma_k$ is the PC scores, $W = V_k$ is the principal directions, and $XW^T = Y_k$.

\subsection{RPCA and Graph Regularized Matrix Factorization}
PCA is sensitive to gross errors. 
RPCA has been proposed to improve it.
Specifically,  RPCA seeks to decompose the data matrix $Y$ into two parts:
\begin{equation}
Y = L + S,
\end{equation}
where $L$ is a low-rank matrix and $S$ is a sparse matrix. The gross errors will be captured by the sparse matrix $S$ and the low-rank matrix $L$ can still approximate $Y$ well. Mathematically, the objective function of RPCA can be written as:
\begin{equation}
\mathop{\min}_{L, S} \norm{L}_* + \rho\norm{S}_1 \ \text{s.t.} \ L+S =Y,
\end{equation}
where $\norm{L}_*$ is called the nuclear norm of $L$, which is the sum of the singular values of $L$. The nuclear norm is a convex relaxation of rank norm.  $ \norm{S}_1$ encourages the sparsity of $S$. However, algorithms dealing with the nuclear norm usually involve computing SVD, which is very time consuming when the problem size is large. To avoid the SVD, one way is to factorize the low-rank matrix $L$ as a product of two low-rank matrices.
For example, Zhou and Tao \cite{zhou2013greedy} formulated a regularized RPCA by letting $L = XW^T$, where $X\in R^{n\times k}$, $W\in R^{p\times k}$ , and $k \ll \min(m, n)$.
Readers may refer to \cite{ma2018efficient} for the recent advances about RPCA.
\par Graph regularized matrix factorization methods consider the local geometrical structure of the data.
For example, Zheng \textit{et al.}  \cite{zheng2011graph} proposed a graph regularized sparse coding method for image presentation:
\begin{equation}
\min \norm{Y - XW^T}_F^2 + \eta \tr(X^T L  X)  + \rho \norm{X}_1,
\end{equation}
where $\eta > 0$, $\rho > 0$. $X$ is the sparse low-dimensional representation of images. $L\in R^{n \times n }$ is the Laplacian matrix of a graph.
Suppose we construct a binary graph matrix $G$ by the $k$-NN approach.
Then the graph regularizer $\tr(X^T L  X) = \frac{1}{2} \sum_{ij} (x_{i\cdot} - x_{j\cdot})^2 G_{ij}$, where $G_{ij}=1$ if sample $i$ is a neighbor of sample $j$ in the constructed graph, otherwise $G_{ij}=0$.
Therefore, it encourages samples that are close in the original space to be  neighbors in the sparse representation $X$. Moreover, Yankelevsky and Elad \cite{yankelevsky2016dual} proposed a low-rank representation method that considers the local geometrical structures in both the feature and sample spaces by imposing two graph regularizers $\tr(X^TL_1 X)$ and $\tr(W^T L_2 W)$.
Here $L_1$ and $L_2$ are the Laplacian matrices in the corresponding spaces, respectively.

\subsection{Probabilistic PCA and its Variants}
\par
Tipping and Bishop \cite{tipping1999probabilistic} showed that PCA could be derived from a Gaussian latent variable model (PPCA).
The generative model of PPCA is:
\begin{equation}
Y=XW^T+E \label{eq:gen_pca},
\end{equation}
where $X\in R^{n\times k }$, $W\in R^{p \times k}$ and $E_{ij} \overset{IID}{\sim} \mathcal{N}(0, \sigma^2)$. The negative log-likelihood is as follows:
\begin{equation}
\frac{1}{2\sigma^2} \norm{Y- XW^T}_F^2 + np \log \sigma. \label{eq:obj_ppca}
\end{equation}
PPCA  conventionally  defines standard Gaussian $\mathcal{N}(0, I)$ prior on each column of $X$ and derives the maximum likelihood estimation (MLE) of $W$ and $\sigma$.

\par  The goal of PPCA is not to give better results than PCA but to permit a broad range of future extensions by facilitating various probabilistic techniques and introducing different assumptions of distribution.
For example, Bishop \cite{bishop1999variational} developed a variational formulation of Bayesian PCA, which can automatically determine the number of retained principal components;
Zhao and Jiang \cite{zhao2006probabilistic} proposed tPPCA which assumes that data are sampled from multivariate Student-$t$ distribution;
Wang \textit{et al.} \cite{wang2012probabilistic} used the Laplace distribution for robust probabilistic matrix factorization.
Most of those methods assume that the underlying distribution of data are IID.
\subsection{PCA with Non-IID Assumption}
\par To allow the model to account for more general noise, Zhao \textit{et al.} \cite{zhao2014robust} relaxed the IID assumption and introduced the mixture of Gaussian noise to model the complex noise. Cao \textit{et al.} further extended the model to the mixture of exponential family noise \cite{Cao_2015_ICCV}.
Note that such noise is not identical but still independent. Inspired by the linear mixed model, Kalaitzis and Lawrence \cite{Kalaitzis2012} proposed RCA to generalize PPCA by adding fixed effects onto Eq. (\ref{eq:gen_pca}), i.e.,
\begin{equation}  \label{eq:RCA}
Y=XW^T + ZV^T + E,
\end{equation}
where $Z$ is a matrix of known covariates. The loading matrices $W$ and $V$ can be marginalized with the Gaussian isotropic priors:
\begin{equation}
\ln p(Y) = \sum_{j=1}^n \ln \mathcal{N}(Y_{\cdot j}| 0, XX^T + \Sigma),
\end{equation}
where $\Sigma = ZZ^T + \sigma^2 I$. $\Sigma$ can be interpreted as an offset of the covariance in the feature space.
There exists a line of research that studies PCA model for data-dependent noise \cite{Han2013, Vaswani2016, Vaswani2017}.
For example, Vaswani and Guo \cite{Vaswani2016} proposed the correlated-PCA model. Given a time sequence of data vectors, $y_t$, that satisfies
\begin{equation}
	y_t = l_t + w_t, \text{ with } w_t = M_t l_t \text{ and } l_t = Pa_t,
\end{equation}  
where $P$ is an $n\times k$ basis matrix, $l_t$ is the true data vector, $w_t$ is the data-independent noise, and $M_t$ is the correlation/data-dependency matrix. 
We note that the noise is only correlated in the feature space.
Gu and Shen \cite{Gu2020} considered the problem that the noise is IID Gaussian but the factor is modeled by a zero-mean Gaussian process.   
\par 
To utilize the covariances in both the sample and feature spaces, Allen \textit{et al} \cite{Allen2014} proposed a generalized PCA (GPCA):
\begin{equation}  \label{eq:gpca}
Y = \sum_{i=1}^{k} d_i x_i w_i^T + E; E\sim \mathcal{MN}(\boldsymbol{0}, \Omega, \Sigma),
\end{equation}
such that $X^T\Omega X=I$, $W^T \Sigma W = I$, and $E$ follows the  matrix normal distribution $\mathcal{MN}_{n, p}(M, \Omega, \Sigma)$.
We note that $\Omega^{-1}$ and $\Sigma^{-1}$ are predefined in GPCA. 
Hence, GPCA can be restrictive for general data.
\par To handle the nonlinearity, some authors extended PPCA inspired by the kernel methods \cite {tipping2001sparse, ge2010kernel, moghaddam2002principal}. Moreover, Lawrence \cite{lawrence2005probabilistic} introduced an alternative probabilistic interpretation of PPCA (DPPCA) and non-linear DPPCA through Gaussian processes.
But those non-linear methods are often difficult to explain.
\par
In tensor data analysis, some researchers also made their efforts to handle comlicated noise in real data \cite{Stegle2011, Ding2014}.
For example, Ding \textit{et al.} \cite{Ding2014} proposed dimension folding PCA (DFPCA) for matrix-valued data:
\begin{equation}
Y = M + XZW^T + E,
\end{equation}
where $X\in R^{n\times k_1}$, $Z\in R^{k_1\times k_2}$, $W\in R^{p\times k_2}$, and $E$ follows the matrix normal distribution.
The observed data are a set of matrices $\{Y_i\}_{i=1}^q$.
Different from GPCA that the precision matrices are predefined, DFPCA can estimate them from data by maximum likelihood estimation.
But it requires the number of observed matrices $q> \max(n/p, p/n)$ + 1 \cite{Dutilleul1999a}. Hence, DFPCA is not applicable when there is only one observed matrix $Y$.
\subsection{Gaussian Graphical Model} \label{sec:sparse_precision}
Gaussian graphical models (GGM) have been proposed to understand the statistical relationship between variables of interest in the form of a graph. Specifically, those models use multivariate Gaussian distribution to model the statistical relationship between variables. The precision matrix of the multivariate Gaussian reveals the conditional correlations between pairs of variables. How to estimate a large precision matrix is fundamental in modern multivariate analysis. Formally, suppose one has $n$ multivariate normal observations of dimension $p$ with covariance $\Sigma$. Let $\Theta = \Sigma^{-1}$ be the precision matrix, and $S$ be the empirical covariance matrix; then the problem is to maximize the log-likelihood:
\begin{equation}
\log |\Theta| -\tr{(S \Theta)} \label{eq:precision_ll},
\end{equation}
where $\Theta$ is positive definite and $|\Theta|$ is the determinant of $\Theta$.

\par To estimate a large precision matrix, the sparse assumption is made in literature, i.e., many entries in the precision matrix $\Theta$ are zeros. Thus, one can add the $L_1$-norm penalty onto the log-likelihood:
\begin{equation}
\log |\Theta| -\tr{(S \Theta)} -\rho \norm{\Theta}_1,
\end{equation}
where $\rho > 0$ controls the sparsity of $\Theta$. This problem has been extensively studied \cite{yuan2007model, friedman2008sparse, li2010inexact, hsieh2014quic, danaher2014joint, cai2016estimating}. Readers may refer to \cite{fan2016overview} for a comprehensive review.

\section{Matrix Normal PCA}
\subsection{Model Construction}
\par Most probabilistic methods assume the underlying noise $E$ is IID. Though the IID assumption enjoys good theoretical properties, it is easily violated in real-world data. To model the underlying correlated noise (graphical noise), a naive approach is to assume that the noise follows a multivariate Gaussian distribution, i.e., $\text{vec}(E)\sim \mathcal{N}(\textbf{0}, \Sigma)$. However, $E$ consists of $np$ variables, the corresponding covariance matrix $\Sigma$ is of size $n^2p^2$ which is too huge and thus infeasible.

\par
\begin{figure}[!t]
    \centering
    \includegraphics[width=0.99\columnwidth]{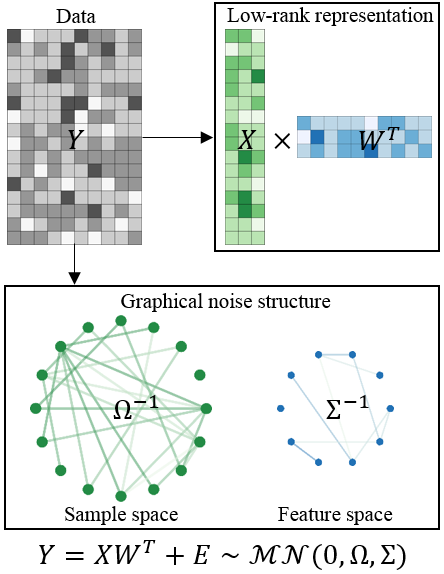}
    \caption{Illustration of MN-PCA. MN-PCA models the low-rank representation and the structure of noise in both feature and sample spaces.}
    \label{fig:model_illtutration}
\end{figure}

 Instead, we assume that the noise is correlated in both the sample and feature spaces. We use a matrix normal distribution to model the among-sample and among-feature  covariances of noise simultaneously. Due to its close relationship with PCA, we name it as MN-PCA (Fig. \ref{fig:model_illtutration}).
Specifically, MN-PCA follows the similar generative model as PPCA
\begin{equation}
Y = XW^T + E.
\end{equation}
But MN-PCA assumes $E \sim \mathcal{MN}_{n,p}(\textbf{0}, \Omega, \Sigma)$, where $\Omega_{n \times n}$ and $\Sigma_{p\times p}$ are among-row and among-column variance respectively. Note that the matrix normal is related to the multivariate normal distribution in the following way:
\begin{equation}
Y \sim \mathcal{MN}_{n, p}(M, \Omega, \Sigma),
\end{equation}
if and only if
\begin{equation}
\text{vec}(Y) \sim \mathcal{N}_{np}(\text{vec}(M), \Sigma \otimes \Omega),
\end{equation}
where $\otimes$ denotes the Kronecker product. Therefore, the density function of $\mathcal{MN}_{n, p}(M, \Omega, \Sigma)$ has the form:
\begin{equation}
P(Y |M, \Omega, \Sigma) = \frac{\exp \left[ -\frac{1}{2} \tr \left( \Sigma^{-1}(Y-M)^T\Omega^{-1}(Y-M)\right) \right]}{(2\pi)^{\frac{np}{2}} |\Omega|^{\frac{n}{2}} |\Sigma|^{\frac{p}{2}} }.
\end{equation}
Accordingly, the negative log-likelihood of MN-PCA is
\begin{align}
\begin{split} \label{eq:likelihood}
\mathcal{L} = & \frac{1}{2} \tr \left[ \Sigma^{-1}(Y-XW^T)^T\Omega^{-1}(Y-XW^T)\right] \\
              & + \frac{p}{2} \log |\Omega| + \frac{n}{2} \log |\Sigma|.
\end{split}
\end{align}
\par The number of parameters of the covariance reduces from $n^2p^2$ to $n^2 + p^2$, but it is still infeasible to minimize the negative log-likelihood in Eq. (\ref{eq:likelihood}).
Because the number of free parameters in $\Omega$ and $\Sigma$  grows quadratically with $n$ and $p$, respectively.
To address this issue, we impose sparse constraints on the precision matrices $\Omega^{-1}$ and $\Sigma^{-1}$ by introducing the $L_1$-norm regularization terms:
\begin{align}
\begin{split} \label{eq:likelihood_reg}
\mathcal{L} = & \frac{1}{2} \tr \left[ \Sigma^{-1}(Y-XW^T)^T\Omega^{-1}(Y-XW^T)\right] \\
& + \frac{p}{2} \log |\Omega| + \frac{n}{2} \log |\Sigma|  + n\lambda_1 \norm{\Omega^{-1}}_1  + p\lambda_2 \norm{\Sigma^{-1}}_1, \\
\end{split}
\end{align}
where $\lambda_1, \lambda_2$ control the sparsity of $\Omega^{-1}$ and $\Sigma^{-1}$, respectively. Note that $\Omega$ and $\Sigma$ are positive-definite. We can further rewrite Eq. (\ref{eq:likelihood_reg}) in the Frobenius norm:
\begin{align}
\begin{split} \label{eq:likelihood_F}
\mathcal{L} = & \frac{1}{2} \norm{\Omega^{- \frac{1}{2} }(Y-XW^T)\Sigma^{- \frac{1}{2} }}_F^2 + \frac{n}{2} \log |\Omega| + \frac{p}{2} \log |\Sigma|\\
& + n\lambda_1 \norm{\Omega^{-1}}_1  + p\lambda_2 \norm{\Sigma^{-1}}_1.
\end{split}
\end{align}
Let $\Omega$ and $\Sigma$ be identity matrices, then $\mathcal{L}$  reduces to
\begin{equation}
\mathcal{L} =  \frac{1}{2} \norm{Y - XW^T}_F^2.
\end{equation}
Therefore, MN-PCA degenerates to the classical PCA when $\Omega$ and $\Sigma$ are identity matrices.

\subsection{Model Interpretation}
\par
\begin{figure}[!t]
    \centering
    \includegraphics[width=\columnwidth]{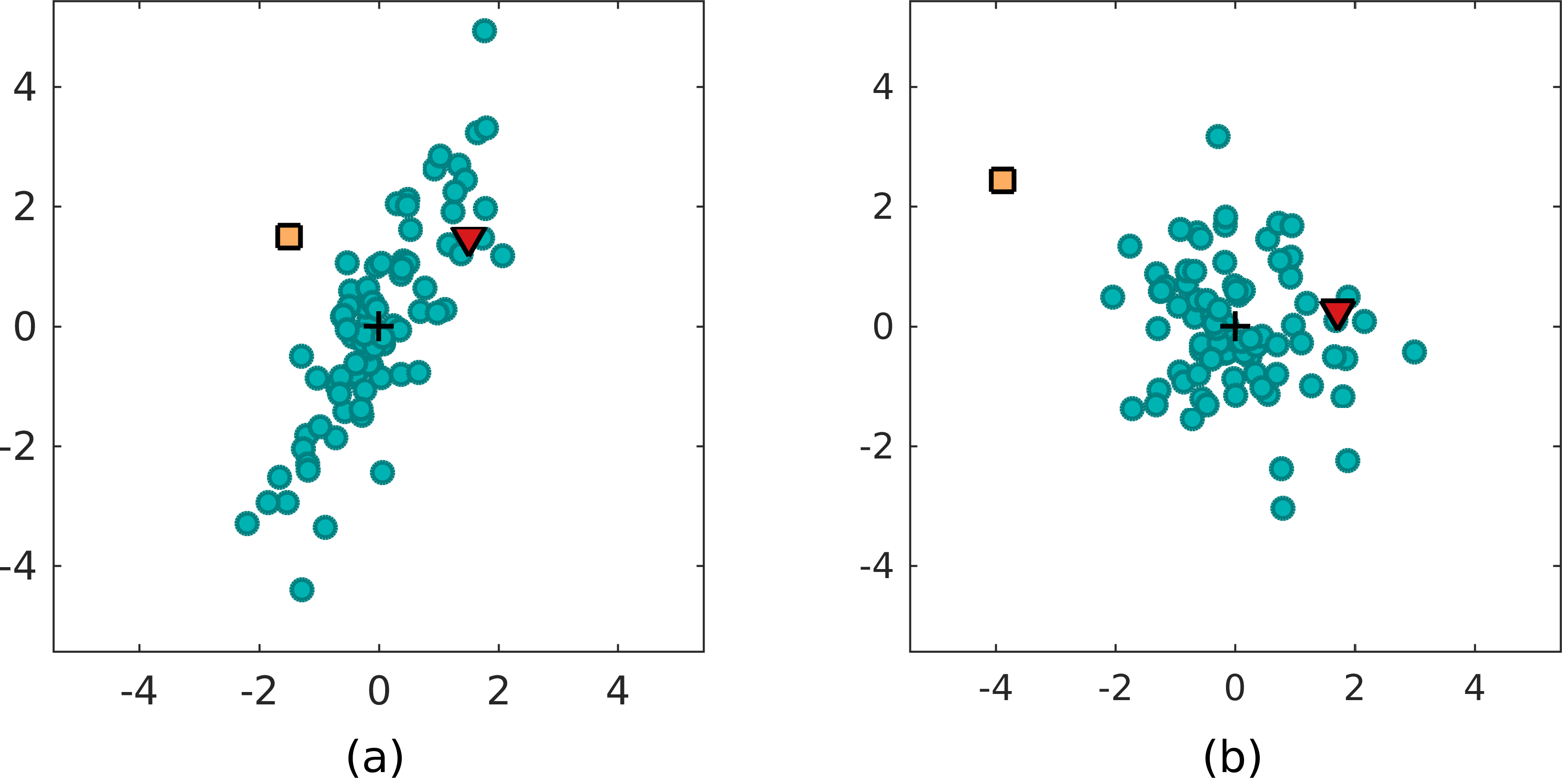}
    \caption{Illustration of the Mahalanobis distance in a 2D data. The blue dots are drawn from a 2D Gaussian distribution. The cross mark indicates the center of the distribution.}
    \label{fig:mdistance}
\end{figure}

PCA aims at minimizing the reconstruction errors, which is measured by the sum of Euclidean distances between the data points $y_{i\cdot}$ and the reconstructed points $x_{i\cdot}W^T$. In particular, the objective function of PCA Eq. (\ref{eq:obj_pca}) can be rewritten as:
\begin{equation}
\min \sum_{i=1}^{n} (y_{i \cdot} - x_{i\cdot}W^T) (y_{i \cdot} - x_{i\cdot}W^T)^T,
\end{equation}
Nevertheless, measuring the distance among points by Euclidean distance can be misleading, knowing that they are realizations of the anisotropic multivariate distribution.
For example, as shown in  Fig. \ref{fig:mdistance}a., there are three points $x_1$ (square), $x_2$ (cross mark) and $x_3$ (triangle) drawn from an anisotropic 2D normal distribution $\mathcal{N}(\mu, \Sigma)$ (the density is illustrated by the blue dots).
$x_2$ is the center of the distribution.
The Euclidean distance between $x_1$ and $x_2$ is equal to that between $x_2$ and $x_3$. This is misleading.
Because the density of the distribution near $x_1$ is much lower
than that near $x_3$.
In a sense, the distance $x_1$ is farther from  $x_2$ than $x_1$.
Hence, Euclidean distance does not reflect the underlying random process $\mathcal{N}(\mu, \Sigma)$.
Mahalanobis distance addresses this problem by utilizing the information of covariance:
\begin{equation}
d(x_i, x_j) = (x_i - x_j)^T\Sigma^{-1} (x_i - x_j),
\end{equation}
which is equivalent to deform the anisotropic Gaussian to an isotropic Gaussian and then compute the Euclidean distance
(Fig. \ref{fig:mdistance}b).
MN-PCA can be regarded as minimizing the reconstruction error over the generalized Mahalanobis distance.
When  $\Omega = I$, the reconstruction error of MN-PCA has the Mahalanobis distance form:
\begin{equation} \label{eq:maha_dist1}
\min \sum_{i=1}^{n} (y_{i \cdot} - x_{i\cdot}W^T)\Sigma^{-1}(y_{i \cdot} - x_{i\cdot}W^T)^T.
\end{equation}
We have the similar result when $\Sigma = I$.
Thus, we can regard MN-PCA as a minimization of the reconstruction error over the generalized Mahalaobis distance.
MN-PCA transforms the matrix normal noise to isotropic noise and then applies PCA to the transformed data.
The following theorem explains the rationale of MN-PCA:
\begin{theorem}
	Suppose $Y=M+E$, where $M\in R^{n\times p}$ is the low-rank structure, and $E\sim \mathcal{MN}(\textbf{0}, \Omega, \Sigma)$ is the matrix normal noise.
	Assume that $Y$ is already centered for simplicity. If $\tr(\Omega)=n\sigma$ and $\tr(\Sigma)=p\sigma$, then the following inequality holds:
	\begin{equation}\label{eq:pca_noise_effect}
	\left|\sigma_i\left(\frac{1}{n}\mathbb{E}(Y^TY)\right) - \sigma_i\left(\frac{1}{n}M^TM \right) \right|\leq \sigma \sigma_1(\Sigma),
	\end{equation}
	where $\sigma_i(A)$ is the $i$-th largest eigenvalue of the matrix $A$.
\end{theorem}
\begin{proof}
	See the Supplementary Materials.
\end{proof}
\noindent
The constraints $\tr(\Omega)=n\sigma$ and $\tr(\Sigma)=p\sigma$  make the scales of IID noise $\mathcal{N}(0,\sigma^2)$ and matrix normal noise $\mathcal{MN}(\textbf{0}, \Omega, \Sigma)$ comparable.
Recall that PCA is interested in the eigenvalues and eigenvectors of the covariance $\frac{1}{n} Y^TY$. Therefore, Eq. (\ref{eq:pca_noise_effect}) quantifies the difference of eigenvalues between the expectation of sample covariance  $\frac{1}{n}Y^TY$ and the noise-free covariance $\frac{1}{n}M^TM$.
Note the upper bound $\sigma \sigma_1(\Sigma)$  is the tightest when the noise is IID, i.e., $\Sigma=\sigma I$.
It implies that PCA is more suitable for IID noise and may be inaccurate when the noise is non-IID.
The result of $\frac{1}{p}YY^T$ can be shown in the same manner.

\subsection{Maximum Regularized Likelihood}
Here we present an alternatively iterative update procedure to seek the maximum regularized likelihood (MRL): 
\begin{align}
\begin{split}
\min               & \mathcal{L} = \frac{1}{2} \tr \left[ \Sigma^{-1}(Y-XW^T)^T\Omega^{-1}(Y-XW^T)\right] \\
& + \frac{p}{2} \log |\Omega| + \frac{n}{2} \log |\Sigma| + n\lambda_1 \norm{\Omega^{-1}}_1  + p\lambda_2 \norm{\Sigma^{-1}}_1\\
\text{s.t. }               &    \Omega \in S^n_{++}, \Sigma \in S^p_{++} ,\\
\end{split}
\end{align}
where $\Omega \in S_n^{++}$ denotes that $\Omega$ is a $n\times n$ symmetric positive definite matrix. This is a matrix optimization problem involving four matrix variables $X$, $W$, $\Omega$ and $\Sigma$. We adopt a block coordinate descent strategy to minimize $\mathcal{L}$. \\
\textbf{Updating $X$ and $W$:} Retain the terms involving $X$ and $W$
\begin{align}
\begin{split}\label{eq:obj_XW}
\min \mathcal{L}  & = \frac{1}{2} \norm{\Omega^{- \frac{1}{2} }Y\Sigma^{- \frac{1}{2} }- \Omega^{- \frac{1}{2} }XW^T\Sigma^{- \frac{1}{2} }}_F^2
\end{split}
\end{align}
Let $Y'$ be $\Omega^{- \frac{1}{2} }Y\Sigma^{- \frac{1}{2} }$. By Eckart and Young's theorem, the best rank-$k$ approximation of $Y'$ in the Frobenius norm is the truncated $k$ SVD of $Y'$, i.e., $Y'_k = U_k \Sigma_k V_k^T$. Let
\begin{equation}
\begin{split} \label{eq:update_XW}
X = \Omega^{\frac{1}{2}} U_k \Sigma_k, \ W=\Sigma^{\frac{1}{2} } V_k.
\end{split}
\end{equation}
Then $\Omega^{- \frac{1}{2} }XW^T\Sigma^{- \frac{1}{2} } = Y'_k$ and the objective function is minimized. Thus we obtain the closed-form solution of $X$ and $W$ for Eq. (\ref{eq:obj_XW}). However, computing $\Omega^{-\frac{1}{2}}$ and $\Sigma^{-\frac{1}{2}}$ is often computationally expensive in practice. Instead, we employ an alternative least square approach (ALS) to update $X$ and $W$
\begin{align}
X = Y\Sigma^{-1}W(W^T\Sigma^{-1}W + \epsilon I)^{-1}
\label{eq:update_X},
\\
W = Y^T \Omega^{-1} X (X^T \Omega^{-1} X + \epsilon I)^{-1}
\label{eq:update_W},
\end{align}
where $\epsilon$ is a small positive constant to protect the inverse of matrices from singular.\\
\textbf{Updating $\Omega$ and $\Sigma$:} A straightforward approach is to optimize $\Omega$ and $\Sigma$ iteratively as follows:
\begin{equation}
\left\{
\begin{aligned} \label{eq:flip_flop}
 \hat{\Omega} = \arg\min _{\Omega} \tr (\Omega^{-1} S_1)  + \log |\Omega| + \lambda_1 \norm{\Omega^{-1}}_1,\\
  \hat{\Sigma} = \arg\min_{\Sigma} \tr (\Sigma^{-1} S_2) + \log|\Sigma|  + \lambda_2 \norm{\Sigma^{-1}}_1,
\end{aligned}
\right.
\end{equation}
where
\begin{align}
S_1 = \frac{1}{p} (Y-XW^T)\Sigma^{-1}(Y-XW^T)^T, \\
S_2 = \frac{1}{n} (Y-XW^T)^T\Omega^{-1}(Y-XW^T).
\end{align}
We can regard $S_1$ and $S_2$ as the empirical covariances, and both problems in Eq. (\ref{eq:flip_flop}) are $L_1$-norm regularized precision matrix estimation, for which efficient optimization has been intensively studied (Sec \ref{sec:sparse_precision}).
We adopt a fast and stable algorithm QUIC \cite{hsieh2014quic} as the basic solver:
\begin{equation}
\left\{
\begin{aligned} \label{eq:flip_flop_l1}
\hat{\Omega} = \text{QUIC}(S_1, \lambda_1), \\
\hat{\Sigma} = \text{QUIC}(S_2, \lambda_2).
\end{aligned}
\right.
\end{equation}

\par The inference scheme is summarized in \textbf{Algorithm} \ref{algo:mn-pca}. Given any invertible matrix $P$, $XPP^{-1}W^T$ equals $XW^T$. Therefore, $X$ and $W$ are not identifiable. To obtain a unique low-rank representation, we use GPCA \cite{Allen2014} to post-process $X$ and $W$. GPCA uses a generalized power method to obtain the columns of $U$ and $V$ sequentially. We note that Eq. (\ref{eq:gpca}) converges to the unique global solution when $\Omega^{-1}$ and $\Sigma^{-1}$ are both positive-definite.
\begin{table}
    \begin{minipage}{\columnwidth}
    \begin{algorithm}[H]
        \begin{algorithmic}[1]\caption{\textbf{Maximum Regularized Likelihood}}\label{algo:mn-pca}
                    \Input data matrix $Y$, rank $k$
                    \Output  $X$, $W$, $\Omega$ and $\Sigma$
                    \State Truncated $k$ SVD of $Y=U_k \Sigma_k V_k$
                    \State Initialize $X=U_k$ and $Y=\Sigma_k V_k$, $A = I$, $\Sigma=I$
                    \Repeat
                        \State update $\Omega$ and $\Sigma$ with Eq. (\ref{eq:flip_flop})
                        \Repeat
                            \State update $X$ with Eq. (\ref{eq:update_X})
                            \State update $W$ with Eq. (\ref{eq:update_W})
                        \Until{convergence}
                    \Until{Change of the objective function value is small enough}
        \end{algorithmic}
    \end{algorithm}
\end{minipage}

\end{table}


\subsection{Minimizing Wasserstein Distance}
\par
MRL is well-understood and easy to implement, yet it is still a point estimation and may get trapped in local minimums that are far from the global minimum.
In the setting of MN-PCA, when the condition number of the among-row variance $\Omega$ or the among-column variance $\Sigma$ is large, the data matrix $Y$ can be significantly influenced by the noise $E$.
So the initial PCA of $Y$ is far from the real low-rank structure of the observed data, and MRL tends to converge to a solution close to the initial PCA. 
To address this issue, we aim to minimize the discrepancy between the white matrix normal distribution and that of the residue instead of maximizing the likelihood function.

\par First, recall the original model:
\begin{equation}
Y \sim \mathcal{MN}_{n, p}(M, \Omega, \Sigma),
\end{equation}
if and only if
\begin{equation}
QYR \sim \mathcal{MN}_{n, p}(QMR, I, I),
\end{equation}
where $Q$ and $R$ are both square, $Q^{T}Q=\Omega^{-1}$, $R^{T}R=B^{-1}$ and $M=XW^T$. It implies that the true factorization of $\Omega^{-1}$ and $\Sigma^{-1}$ transform the residue matrix to IID Gaussian noise. Then we transform the objective function by defining a divergence $D(\cdot||\cdot)$:
\begin{equation}
\min_{Q,R} D(Q(Y-M)R||\mathcal{MN}_{n, p}(\textbf{0}_{n\times p}, I_{n}, I_{p})).
\end{equation}
The Wasserstein distance \cite{Villani2008} is a good metric of measuring the divergence between two distributions.
If the Wasserstein distance is minimized to 0, the density functions of the two distributions are the same except a set with measure 0.
Compared with MRL, minimizing the Wasserstein distance has the potential to find a solution closer to the global optimum.
The definition of Wasserstein distance is
\begin{equation}
W_c(P_x, P_y) = \inf_T \mathbb{E}_{x\sim p_x}[c(x, T(x))],
\end{equation}
where $T: X\rightarrow Y$ is a measure preserving map. The Wasserstein distance is hard to compute due to the infimum operator. However, if both $P_x$ and $P_y$ are Gaussian distributions, their Wasserstein distance has a closed form solution.
Given the two normal distributions $x=\mathcal{N}(m_{1},\Sigma_{1})$ and $y=\mathcal{N}(m_{2},\Sigma_{2})$ with the means $m_1$, $m_2\in R^{p}$ and the covariance $\Sigma_1$, $\Sigma_2 \in R^{p\times p}$, their squared Wasserstein distance is defined as \cite{olkin1982distance}:
\begin{equation}
\label{eq:W2}
W(x,y)_{2}^2=\|m_{1}-m_{2}\|^{2}_{2}+\tr\left(\Sigma_{1}+\Sigma_{2}-2\left(\Sigma^{\frac{1}{2}}_{2}\Sigma_{1}\Sigma^{\frac{1}{2}}_{2}\right)^{\frac{1}{2}}\right).
\end{equation}
We use $E$ to denote the current residue, $E  = Y-M$, $\tilde{E}$ to denote $QER$, and $\sigma ^{2}$ to denote the variance of noise. To pursue sparsity, we introduce the $L_{1}$-norm regularization as before. Note that the expectation of mean of $E$ is zero, so we omit the computation of mean for both the norm and covariance in Eq. (\ref{eq:W2}).
When we update $Q$, the estimation of covariance among rows should be:
\begin{equation}
\Sigma=\frac{1}{p}\tilde{E}\tilde{E}^{T},
\end{equation}
then the objective function turns to:
\begin{equation}
\min_{Q}\tr\left(\Sigma -2\sigma \Sigma^{\frac{1}{2}}\right)+ \lambda_{1}\|Q^{T}Q\|_{1},
\end{equation}
The formula can be simpler since
\begin{equation}
\tr(\Sigma)=\frac{1}{p}\tr(\tilde{E}\tilde{E}^{T})=\frac{1}{p}\|\tilde{E}\tilde{E}^{T}\|_2^2,
\end{equation}
and
\begin{equation}
\tr(\Sigma^\frac{1}{2})=\frac{1}{\sqrt{p}}\tr(\tilde{E}\tilde{E}^{T})^{\frac{1}{2}}=\frac{1}{\sqrt{p}}\|\tilde{E}\|_{*}.
\end{equation}
Hence, the objective function to update $Q$ takes the form:
\begin{equation}\label{eq:Q_update}
\min_{Q}\frac{1}{p}\|QER\|_{2}^{2}-\frac{2\sigma}{\sqrt{p}}\|QER\|_{*}+ \lambda_{1}\|Q^{T}Q\|_{1}.
\end{equation}
Similarly, the objective function to update $R$ is:
\begin{equation}\label{eq:R_update}
\min_{R}\frac{1}{n}\|QER\|_{2}^{2}-\frac{2\sigma}{\sqrt{n}}\|QER\|_{*}+ \lambda_{2}\|R^{T}R\|_{1}.
\end{equation}
In Eq. (\ref{eq:Q_update}), the coefficient ratio of terms $||QER||^2_2$ and $||QER||_*$ is $1/2\sigma\sqrt{p}$ and in Eq. (\ref{eq:R_update}), the ratio is $1/2\sigma\sqrt{n}$.
If we update $Q$ and $R$ through Eqs. (\ref{eq:Q_update}) and (\ref{eq:R_update}) alternatively, the optimization will not converge to the solution, since the coefficient ratios are not consistent.
To address this issue, we unify the coefficients in Eqs. (\ref{eq:Q_update}) and (\ref{eq:R_update}) with their geometric average $1/\sqrt{np}$ and $2\sigma /\sqrt[4]{np}$. Then the objective function is changed to:
\begin{equation}\label{eq:w2_relax}
\min_{Q,R}\frac{1}{\sqrt{np}}\|QER\|_{2}^{2}-\frac{2\sigma}{\sqrt[4]{np}}\|QER\|_{*}+ \lambda_{1}\|Q^{T}Q\|_{1}+ \lambda_{2}\|R^{T}R\|_{1}.
\end{equation}
Not that if
\begin{equation}
Q^{*},R^{*}=\mathop{\arg\min}_{Q,R}\frac{1}{\sqrt{np}}\|QER\|_{2}^{2}-\frac{2}{\sqrt[4]{np}}\|QER\|_{*},
\end{equation}
then
\begin{equation}
\sqrt{\sigma}Q^{*},\sqrt{\sigma}R^{*}=\mathop{\arg\min}_{Q,R}\frac{1}{\sqrt{np}}\|QER\|_{2}^{2}-\frac{2\sigma}{\sqrt[4]{np}}\|QER\|_{*}.
\end{equation}
Thus, there is no need to adjust the parameter $\sigma$. In the experiments, we find that the optimizing process converges faster with smaller $\sigma$. To keep the balance between $Q$ and $R$, we normalize $Q$ after each iteration, i.e., the average eigenvalues of $Q$ and $R$ are equal. Besides, the first term of the relaxed objective function Eq. (\ref{eq:w2_relax}) is quadratic to $Q$ and $R$, which can be solved efficiently.
To compute the gradient with respect to $Q$ and $R$ through Eq. (\ref{eq:w2_relax}), we need to know the gradient of the nuclear norm $||QER||_*$.
Since the nuclear norm is non-derivable, we can only obtain the subgradient. If the SVD of $QER$ is $U_0\Sigma_0V_0^T$, then a subderivative is $U_0V_0^T$ \cite{Watson1992}. Then the subgradients of $Q$ and $R$ through Eq. (\ref{eq:w2_relax}) is:

\begin{align}
\nabla_Q = &\frac{2}{\sqrt{np}}QERR^TE^T - \frac{2}{\sqrt[4]{np}}U_0V_0^TR^TE^T \\\nonumber
&+ \lambda_{1}\text{sign}(Q^TQ)Q, \\
\nabla_R =& \frac{2}{\sqrt{np}}E^TQ^TQER - \frac{2}{\sqrt[4]{np}}E^TQ^TU_0V_0^T \\\nonumber
&+ \lambda_{2}\text{sign}(R^TR)R.
\end{align}
Then $Q$ and $R$ are updated by:
\begin{align}
Q^+ = Q-\eta\nabla_Q, \label{eq:update_Q}\\
R^+ = R-\eta\nabla_R, \label{eq:update_R}
\end{align}
where $\eta$ is the learning rate.
The use of Wasserstein metric is inspired from generative adversarial network (GAN) \cite{Goodfellow2014},  especially the Wasserstein-GAN \cite{Gulrajani2017}.
Note that if we treat $Q$ and $R$ as the weight matrices of a layer of neural network, the data matrix as input and the truncated SVD as the activate function over the eigenvalue of $Y$, our MN-PCA model can be regarded as a single layer network.
In Wasserstein GAN, the Wasserstein distance between the generated distribution and the target distribution is estimated by a discriminator, while we can directly compute the Wasserstein distance with a close form formula in MN-PCA.
We implement the algorithm by Pytorch with the Adam optimizer.
The gradients can be computed automatically.

\par If the likelihood function is maximized, the residue may not be Gaussian. In contrast, if the Wasserstein distance of two distributions is minimized, then the two distributions are the same almost everywhere. In MN-PCA, theoretically, we have
\begin{theorem}\label{th:W2}
	Suppose $E$ and $R$ are fixed and $n>p$. Let $row\{QER\}$ denote the set of rows of the matrix $QER$, $S$ is the set of $n$ samples from the distribution $N(\textbf{0},I_p)$, $H$ is a $n\times p$ matrix by stacking each item in $S$ as a row in a certain order. Then $W(row\{QER\},S)$ is minimized if and only if $QER$ can be transformed to $H_{p-k}$ by rearranging its rows, where $p-k$ is the rank of $E$ and $H_{p-k}$ is the truncated $p-k$ SVD of $H$.
\end{theorem}
\begin{proof}
	See the Supplementary Materials.
\end{proof}
\par
Theorem \ref{th:W2} shows that under moderate assumptions, the minimum of Wasserstein distance corresponds to the global optimum of the problem.
However, computing the Wasserstein distance exactly is prohibitive.
Thus, we adopt the formula in Eq. (\ref{eq:W2}) instead.
The formula in Eq. (\ref{eq:W2}) is the Wasserstein distance of two multi-variate Gaussian.
However, the real distribution of the residue is unknown. By utilizing the formula, we actually approximated the distribution of residue with its mean vector and covariance matrix.
Thus, the computation of Wasserstein distance is inaccurate.
Since our model is over-parameterized, if we enforce the residue to be exactly IID Gaussian, the model can be overfitted. Thus, the inaccuracy in Eq. (\ref{eq:W2}) makes it possible to adjust our model to different kinds of data.  Moreover, the computation of the formula doesn't need sampling from the distribution of IID Gaussian, which is also intractable.
\begin{table}
    \begin{minipage}{\columnwidth}
        \begin{algorithm}[H]
            \begin{algorithmic}[2]\caption{\textbf{MN-PCA-W2}}\label{alg:W2}
                \Input data matrix $Y$, rank $k$, $\lambda_{1}$, $\lambda_{2}$
                \Output  $X$, $Q$ and $R$
                \Repeat
                \State Truncated $k$ SVD of $X=(QYR)_{k}$
                \State set $E=Y-Q^{-1}XR^{-1}$
                \State update $Q$ with Eq. (\ref{eq:update_Q})
                \State update $R$ with Eq. (\ref{eq:update_R})
                \Until{convergence}
            \end{algorithmic}
        \end{algorithm}
    \end{minipage}
\end{table}

\subsection{Computational Complexity and Convergence}\label{sec:complexity}
\par 
Here we first discuss the computational complexity of MLR. At each iteration of updating $X$, the computational cost arises in matrix multiplication and inverse, which is $O(np^2 + k^3)$. Similarly, the computational cost for updating $X$ is $O(n^2p + k^3)$.
\par Solving $\Omega^{-1}$ and $\Sigma^{-1}$ is much more computationally expensive. The dominant computational cost of QUIC is using the Cholesky decomposition at each iteration for the linear search of the step size to ensure the precision matrix is positive-definiteness. The the complexity of the Cholesky decomposition is $O(n^3)$ for updating $\Omega^{-1}$ ($O(p^3)$ for updating $\Sigma^{-1}$). Therefore, updating $\Omega^{-1}$ and $\Sigma^{-1}$ can be very time-consuming. To address this problem, QUIC decomposes the precision matrix into smaller blocks with connected components and then runs the Cholesky decomposition for each block.
For example, if $\Omega^{-1}$ consists of $C$ connected components of size $n_1, \cdots, n_C$ ($n_1 + \cdots + n_C = n$). Then the time complexity of the Cholesky decomposition is reduced to $O(\max(\{n_i^3\}_{i = 1}^C ))$.
The connected components of $\Omega^{-1}$ can be detected in $O(\norm{\Omega^{-1}}_0)$, which is very efficient when $\Omega^{-1}$ is sparse (the situation of $\Sigma^{-1}$ is the same).
The implementation of QUIC also exploits the sparsity of $\Omega^{-1}$ and $\Sigma^{-1}$ by using sparse matrix operations.
The computational cost of QUIC reduces sufficiently when the estimated precision matrix is sparse.
Empirically, the computational cost is affordable with sufficient large $\lambda_1$ and $\lambda_2$ when $\max(n, p) < 10^4$.
MRL can be regarded as a  block coordinate descent algorithm. 
We note that updating $X$ and $W$ has the closed-form solution when $\Omega$ and $\Sigma$ are fixed. 
However, it is difficult to obtain a global convergence rate for updating $\Omega$ and $\Sigma$. 
Empirically, the objective function value of MRL gets stable in less than ten rounds of iterations.
\par
The complexity in each iteration of MN-PCA by minimizing the Wasserstein distance mainly arises in SVD, which is $O(\mbox{min}(pn^2,np^2))$ and the inverse of $Q$ and $R$, which are $O(n^3)$ and $O(p^3)$, respectively.
Since we only update $Q$ and $R$ with one gradient descent step, MN-PCA-W2 needs more iterations and usually more time than MRL.

\begin{table*}[hpbt]
	\centering
	\begin{threeparttable}
		\caption{Performance of low-rank recovery on Small Synthetic Datasets}
		\label{table_syn_lowrank}
		\begin{tabular}{lccccccccc}
			\toprule
			& \multicolumn{3}{c}{PSNR} & \multicolumn{3}{c}{RMSE} &  \multicolumn{3}{c}{NMI}\\
			\cmidrule(lr){2-4} \cmidrule(lr){5-7} \cmidrule(lr){8-10}
			& PCA &MN-PCA &MN-PCA-w2  &PCA &MN-PCA &MN-PCA-w2&   PCA &MN-PCA& MN-PCA-w2 \\
			\midrule
			$c=8$   & 15.71(0.24) & \textbf{16.79(0.19)} & 16.64(0.17)          & 0.16(0.00) & \textbf{0.14(0.00)} & 0.15(0.00)          & 98.93(1.26)  & \textbf{99.31(0.89)} & 98.22(1.69)           \\
			$c=16$  & 14.15(0.72) & \textbf{16.34(0.28)} & 16.28(0.23)          & 0.20(0.02) & \textbf{0.15(0.00)} & 0.15(0.00)          & 98.19(1.93)  & \textbf{99.29(0.91)} & 99.30(0.90)           \\
			$c=32$  & 9.21(1.48)  & \textbf{15.64(0.36)} & 13.12(3.55)          & 0.35(0.05) & \textbf{0.17(0.01)} & 0.24(0.10)          & 73.87(12.75) & \textbf{97.94(1.75)} & 82.88(18.82)          \\
			$c=64$  & 5.62(1.24)  & 11.07(4.16)          & \textbf{11.36(3.34)} & 0.53(0.08) & 0.31(0.15)          & \textbf{0.29(0.10)} & 40.33(12.43) & 70.00(34.42)         & \textbf{75.40(19.51)} \\
			$c=96$  & 3.09(1.49)  & 6.94(3.71)           & \textbf{10.37(3.02)} & 0.71(0.12) & 0.48(0.15)          & \textbf{0.32(0.09)} & 17.89(14.10) & 43.42(36.95)         & \textbf{70.16(18.95)} \\
			$c=128$ & 1.52(1.24)  & 4.32(1.09)           & \textbf{9.58(3.40)}  & 0.85(0.12) & 0.61(0.07)          & \textbf{0.35(0.12)} & 7.33(6.33)   & 25.62(27.49)         & \textbf{58.78(34.88)} \\
			$c=160$ & 0.43(1.14)  & 2.83(1.92)           & \textbf{9.33(3.19)}  & 0.96(0.13) & 0.74(0.17)          & \textbf{0.36(0.12)} & 3.68(2.21)   & 18.13(25.01)         & \textbf{59.25(34.74)} \\
			$c=192$ & -0.42(1.18) & 1.78(1.94)           & \textbf{9.62(3.24)}  & 1.06(0.14) & 0.83(0.19)          & \textbf{0.35(0.14)} & 2.68(1.41)   & 17.39(25.24)         & \textbf{68.71(28.93)} \\
			$c=224$ & -1.14(1.24) & 0.54(1.80)           & \textbf{8.99(3.28)}  & 1.15(0.16) & 0.96(0.20)          & \textbf{0.38(0.14)} & 2.08(1.07)   & 13.00(19.45)         & \textbf{63.86(27.14)}\\
			\bottomrule
		\end{tabular}\label{table:syn_small_recovery}
	\end{threeparttable}
\end{table*}

\subsection{Tuning Parameters Selection}
\par Given the desired dimension $k$, we should select the tuning parameters $\lambda_1$ and $\lambda_2$ to balance the trade-off of the log-likelihood value and the sparseness of the inferred precision matrices.
While carrying a 2D grid search of $\lambda_1$ and $\lambda_2$ is computational expensive, we select $\lambda_1$ and $\lambda_2$ separately.
Specifically, we use the Bayesian Information Criterion (BIC) to select appropriate $\lambda_1$. The BIC for the $L_1$-norm regularized precision matrix for a fixed value of $\lambda_1$ is given in \cite{yuan2007model}:
\begin{equation}\label{eq:BIC}
\text{BIC}(\lambda_1) = -\log |\Omega^{-1}(\lambda_1)| + \tr (S_1 \Omega^{-1}(\lambda_1)) + t_1 \frac{\log p}{p},
\end{equation}
where $\Omega^{-1}(\lambda_1)$ is inferred by Eq. (\ref{eq:flip_flop_l1}). $X$ and $W$ are given by truncated $k$ SVD and $\Sigma^{-1}=I$. $t_1$ is the number of non-zero entries in the upper diagonal portion of $\Omega^{-1}(\lambda_1)$.
\par To obtain a grid of values of $\lambda_1$, we use the heuristic approach proposed in \cite{zhao2012huge}. The largest value of the candidate $\lambda_1$ depends on the value of the empirical covariance:
\begin{equation}
\lambda_{\max} = \max\left(\max (S_1 - I)_{ij}  - \min (S_1 - I)_{ij} \right)_{ij}.
\end{equation}
We set $\lambda_{\min} = 0.1\lambda_{\max}$ and the candidate tuning parameters are ten values logarithmically spaced between $\lambda_{\min}$ and $\lambda_{\max}$. The value of $\lambda_1$ minimizing the BIC is chosen as the final tuning parameter.
$\lambda_2$ is chosen by the same procedure.
For MN-PCA-w2, we empirically set $\lambda_{1}=\lambda_{2}=0.05$ in all the experiments.

\section{Experimental Results}
\par
In this section, we demonstrate the effectiveness of the proposed methods on both synthetic data and real-world data. We first compare MN-PCA and PCA on both small-scale and large-scale synthetic data sets and then apply MN-PCA to non-linear synthetic data. We further conduct extensive experiments on various real-world data. All experiments are performed on a desktop computer with a 2GHz Intel Xeon E5-2683v3 CPU, a GTX 1080 GPU card and 16GB memory.
\subsection{Synthetic Experiments}
\subsubsection{Small-Scale Synthetic Data}
\par We construct the toy data with $Y = M + E$, where $Y\in R^{n\times p}$ is the observed data matrix, $M$ is the low-rank signal and $E$ is the matrix normal noise. Each row of $M$ are  drawn from $k$ different centroids which are row vectors of length $p$. We then add the matrix normal noise $E\sim \mathcal{MN}(\textbf{0}, \Omega, \Sigma)$ to $M$.
As we assume the precision matrix $\Omega^{-1}$ is sparse, we use the  sparsity parameters $\alpha_1$ and condition number $c_1$ to control the generated $\Omega^{-1}$.
In particular, the sparsity parameter indicates the proportion of nonzero entries in  $\Omega^{-1}$, $\alpha_1=\# \{\Omega^{-1} \neq 0 \}/n^2$; the condition number  is defined as $c_1=\sigma_{\text{max}}(\Omega^{-1})/\sigma_{\text{min}}(\Omega^{-1})\geq 1$, where $\sigma_{\text{max}}(\Omega^{-1})$ and $\sigma_{\text{min}}(\Omega^{-1})$ are the largest and smallest singular values of matrix $\Omega^{-1}$, respectively.
Note that the larger the condition number is, the violation of the IID assumption is more serious.
Similarly, we generate $\Sigma^{-1}$ with parameters $\alpha_2$ and $c_2$. We always set $c_1 = c_2=c$, $\alpha_1 = \alpha_2= \alpha$ for simplicity in the following experiments. Note that $c=1$ and $\alpha=0$ correspond to the IID situation.

\par We generate data with $n=300$, $p=200$, $k=2$, $\alpha=10^{-2}$. $M$ consists of 300 samples drawn evenly from three different centroids. Specifically, the first and last 20 entries of centroid 1 equal one; the first and last 20 entries of centroid 2 equal minus one; the first 20 entries of centroid 3 equal one and the last 20 entries equal minus one. To investigate the effects of the graphical noise, we vary the condition number $c$ form 8 to 224.
We generate 10 synthetic data for each parameter combination. The $L_1$-norm penalty parameters $\lambda_1$, $\lambda_2$ of the maximum likelihood are selected by the aforementioned procedure.

\par We investigate the effect of graphical noise on the performance of PCA and the proposed methods measured in terms of low-rank matrix recovery (Table \ref{table:syn_small_recovery}). Specifically, let the estimated low-rank matrix be $\hat{M}$, and the true low-rank matrix be $M^*$. We calculate the root mean square root (RMSE) by  $\norm{\hat{M} - M^{*}}_F/\sqrt{np}$ and peak signal to noise ratio (PSNR), respectively. We also apply K-means clustering to the projected data and compute the normalized mutual information (NMI) to evaluate the low-rank recovery implicitly.
Both the proposed methods outperform PCA in terms of all computed metrics. PCA works well when $c=8$ because the distribution of the noise is close to a IID normal distribution when $c$ is small. The performance of the three methods decreases as the condition numbers increase.
The maximum likelihood method becomes unstable along with the increase of the condition numbers.
On the other hand, MN-PCA-w2 is more robust and outperforms MN-PCA when the condition numbers are sufficiently large.

\par We compare the performance of the proposed methods with QUIC in terms of the estimation of precision matrices (Table \ref{table:syn_small_network}). To facilitate the comparison, we apply QUIC to among-row and among-column covariance, respectively. Since there are many small values in the true precision matrices, we only treat the top 150 non-diagonal entries as non-zero values (by absolute value). We summarize the performance in terms of true positive rate (TPR), true negative rate (TNR) and predictive positive value (PPV) (Supplementary Materials).

\begin{table*}[hpbt]
	\centering
	\begin{threeparttable}
		\caption{Performance of precision matrix estimation on Small Synthetic Datasets}
		\label{table:syn_small_network}
		\begin{tabular}{lccccccccc}
			\toprule
			& \multicolumn{3}{c}{$\text{TPR}_1$} & \multicolumn{3}{c}{$\text{TNR}_1$} & \multicolumn{3}{c}{$\text{PPV}_1$}\\
			\cmidrule(lr){2-4} \cmidrule(lr){5-7}  \cmidrule(lr){5-7} \cmidrule(lr){8-10}
			& QUIC &MN-PCA &MN-PCA-w2 & QUIC &MN-PCA & MN-PCA-w2& QUIC &MN-PCA & MN-PCA-w2\\
			\midrule
			$c=8$   & 0.17(0.05) & \textbf{0.35(0.08)} & 0.12(0.03)          & 1.00(0.00) & 1.00(0.00) & 1.00(0.00) & 0.58(0.12) & \textbf{0.75(0.12)} & 0.19(0.04) \\
			$c=16$  & 0.21(0.06) & \textbf{0.45(0.11)} & 0.18(0.03)          & 1.00(0.00) & 1.00(0.00) & 1.00(0.00) & 0.33(0.08) & \textbf{0.73(0.16)} & 0.26(0.04) \\
			$c=32$  & 0.25(0.06) & \textbf{0.45(0.10)} & 0.22(0.04)          & 1.00(0.00) & 1.00(0.00) & 1.00(0.00) & 0.33(0.05) & \textbf{0.80(0.15)} & 0.31(0.03) \\
			$c=64$  & 0.27(0.05) & \textbf{0.38(0.09)} & 0.26(0.03)          & 1.00(0.00) & 1.00(0.00) & 1.00(0.00) & 0.34(0.05) & \textbf{0.83(0.13)} & 0.34(0.03) \\
			$c=96$  & 0.28(0.05) & \textbf{0.39(0.09)} & 0.29(0.03)          & 1.00(0.00) & 1.00(0.00) & 1.00(0.00) & 0.35(0.05) & \textbf{0.72(0.18)} & 0.37(0.03) \\
			$c=128$ & 0.30(0.05) & \textbf{0.38(0.10)} & 0.30(0.03)          & 1.00(0.00) & 1.00(0.00) & 1.00(0.00) & 0.37(0.04) & \textbf{0.75(0.16)} & 0.38(0.02) \\
			$c=160$ & 0.29(0.05) & \textbf{0.34(0.09)} & 0.32(0.04)          & 1.00(0.00) & 1.00(0.00) & 1.00(0.00) & 0.37(0.04) & \textbf{0.78(0.15)} & 0.39(0.03) \\
			$c=192$ & 0.31(0.05) & \textbf{0.33(0.09)} & 0.32(0.04)          & 1.00(0.00) & 1.00(0.00) & 1.00(0.00) & 0.38(0.04) & \textbf{0.76(0.16)} & 0.39(0.03) \\
			$c=224$ & 0.30(0.06) & 0.31(0.09)          & \textbf{0.32(0.04)} & 1.00(0.00) & 1.00(0.00) & 1.00(0.00) & 0.38(0.05) & \textbf{0.74(0.17)} & 0.39(0.03)\\
			\midrule
			& \multicolumn{3}{c}{$\text{TPR}_2$} & \multicolumn{3}{c}{$\text{TNR}_2$} & \multicolumn{3}{c}{$\text{PPV}_2$}\\
			\cmidrule(lr){2-4} \cmidrule(lr){5-7}  \cmidrule(lr){5-7} \cmidrule(lr){8-10}
			$c=8$   & 0.09(0.02) & \textbf{0.22(0.05)} & 0.17(0.03)          & 0.99(0.00) & 1.00(0.00) & 0.99(0.00) & 0.12(0.03) & \textbf{0.94(0.05)} & 0.20(0.02) \\
			$c=16$  & 0.11(0.03) & \textbf{0.30(0.06)} & 0.21(0.05)          & 0.99(0.00) & 1.00(0.00) & 0.99(0.00) & 0.14(0.03) & \textbf{0.84(0.10)} & 0.24(0.04) \\
			$c=32$  & 0.13(0.03) & \textbf{0.28(0.07)} & 0.24(0.05)          & 0.99(0.00) & 1.00(0.00) & 0.99(0.00) & 0.16(0.03) & \textbf{0.84(0.11)} & 0.26(0.03) \\
			$c=64$  & 0.14(0.03) & 0.19(0.06)          & \textbf{0.26(0.06)} & 0.99(0.00) & 1.00(0.00) & 0.99(0.00) & 0.18(0.03) & \textbf{0.82(0.13)} & 0.28(0.04) \\
			$c=96$  & 0.15(0.03) & 0.22(0.06)          & \textbf{0.28(0.08)} & 0.99(0.00) & 1.00(0.00) & 0.99(0.00) & 0.18(0.03) & \textbf{0.65(0.14)} & 0.29(0.05) \\
			$c=128$ & 0.15(0.03) & 0.19(0.05)          & \textbf{0.28(0.07)} & 0.99(0.00) & 1.00(0.00) & 0.99(0.00) & 0.18(0.03) & \textbf{0.68(0.14)} & 0.28(0.05) \\
			$c=160$ & 0.15(0.03) & 0.16(0.05)          & \textbf{0.28(0.08)} & 0.99(0.00) & 1.00(0.00) & 0.99(0.00) & 0.18(0.03) & \textbf{0.70(0.13)} & 0.28(0.05) \\
			$c=192$ & 0.14(0.04) & 0.16(0.04)          & \textbf{0.30(0.07)} & 0.99(0.00) & 1.00(0.00) & 0.99(0.00) & 0.18(0.04) & \textbf{0.72(0.16)} & 0.30(0.05) \\
			$c=224$ & 0.15(0.03) & 0.14(0.05)          & \textbf{0.30(0.07)} & 0.99(0.00) & 1.00(0.00) & 0.99(0.00) & 0.18(0.03) & \textbf{0.65(0.16)} & 0.30(0.05)\\
			\bottomrule
		\end{tabular}
		\begin{tablenotes}
			\small
			\item $\text{TPR}_1$, $\text{TNR}_1$, $\text{PPV}_1$ denote true positive rate, true negative rate and positive predictive value of estimated $\Omega^{-1}$, respectively. $\text{TPR}_2$, $\text{TNR}_2$, $\text{PPV}_2$ denote the corresponding metrics of estimated $\Sigma^{-1}$, respectively.
		\end{tablenotes}
	\end{threeparttable}
\end{table*}

\begin{table*}[hpbt]
    \caption{Performance of low-rank recovery on Large Synthetic Datasets}
    \centering
    \begin{tabular}{lccccccccc}
        \toprule
        & \multicolumn{3}{c}{PSNR} & \multicolumn{3}{c}{RMSE} &  \multicolumn{3}{c}{NMI}\\
        \cmidrule(lr){2-4} \cmidrule(lr){5-7} \cmidrule(lr){8-10}
        & PCA &MN-PCA &MN-PCA-w2  &PCA &MN-PCA &MN-PCA-w2&   PCA &MN-PCA& MN-PCA-w2 \\
        \midrule
 $p=2000$&4.53(0.50)& 20.06(1.19) &  \textbf{20.99(0.17)}&0.59(0.03)& 0.10(0.01) & \textbf{ 0.09(0.00)}&4.32(7.86)& \textbf{98.80(1.18)} &  96.70(2.17)\\
 $p=3000$&4.79(0.46)& 7.49(0.65) &  \textbf{20.59(0.23)}&0.58(0.03)& 0.42(0.03) &  \textbf{0.09(0.00)}&4.88(7.34)& 85.13(9.73) &  \textbf{96.75(1.80)}\\
 $p=4000$&4.10(0.51)& 12.51(5.74) &  \textbf{14.96(0.97)}&0.62(0.04)& 0.28(0.17) &  \textbf{0.18(0.02)}&0.83(0.09)& \textbf{90.75(11.66)} &  73.47(7.52)\\
 $p=5000$&4.47(0.25)& 7.76(0.38) &  \textbf{10.87(0.22)}&0.60(0.02)& 0.41(0.02) &  \textbf{0.29(0.01)}&1.42(0.44)& \textbf{82.98(11.91)} &  47.62(1.57)\\
 $p=6000$&3.71(0.66)& 5.60(2.30) &  \textbf{10.69(0.20)}&0.65(0.05)& 0.54(0.14) &  \textbf{0.29(0.01)}&1.57(0.53)& \textbf{59.43(32.38)} &  44.40(1.65)\\
        \bottomrule
    \end{tabular}\label{table:large_lr}
\end{table*}

\par Numerical results from Table  \ref{table:syn_small_network} reveal some interesting conclusions.
1) MN-PCA outperforms QUIC and MN-PCA-w2.
2) The TPR of all methods is relatively low. It is difficult to recover all non-zero entries in the true precision matrices of the noise as there may exist many small values.
3) The TNR of all methods is close to one. The BIC criterion is prone to choose sparser models, as discussed in \cite{zhao2012huge, danaher2014joint}.
4) It is not surprising that QUIC has lower TPR due to the existence of the low-rank matrix. But the TPR of QUIC increases along with the increase of the condition numbers. It is easier for QUIC to find true interactions when the effect of the matrix normal noise is stronger.
5) On the other hand, the TPR of the proposed methods increases first and then decreases with the condition numbers increase. When the effect of the matrix normal noise is small, it is difficult to capture the structure of the precision matrices. However, if the effect of the noise is too strong, the estimated low-rank matrix may mislead the estimation of precision matrices.
6) The PPV of MN-PCA is significantly higher than that of QUIC. Therefore, the FDR ($\text{FDR} = 1 - \text{PPV}$) of the proposed methods is acceptable when the condition number is mild. MN-PCA may help scientists to discover the underlying statistical correlations in the noise, which are ignored by PCA and most of its variants.


\par
\begin{figure*}[!t]
    \centering
    \includegraphics[width=1.5\columnwidth]{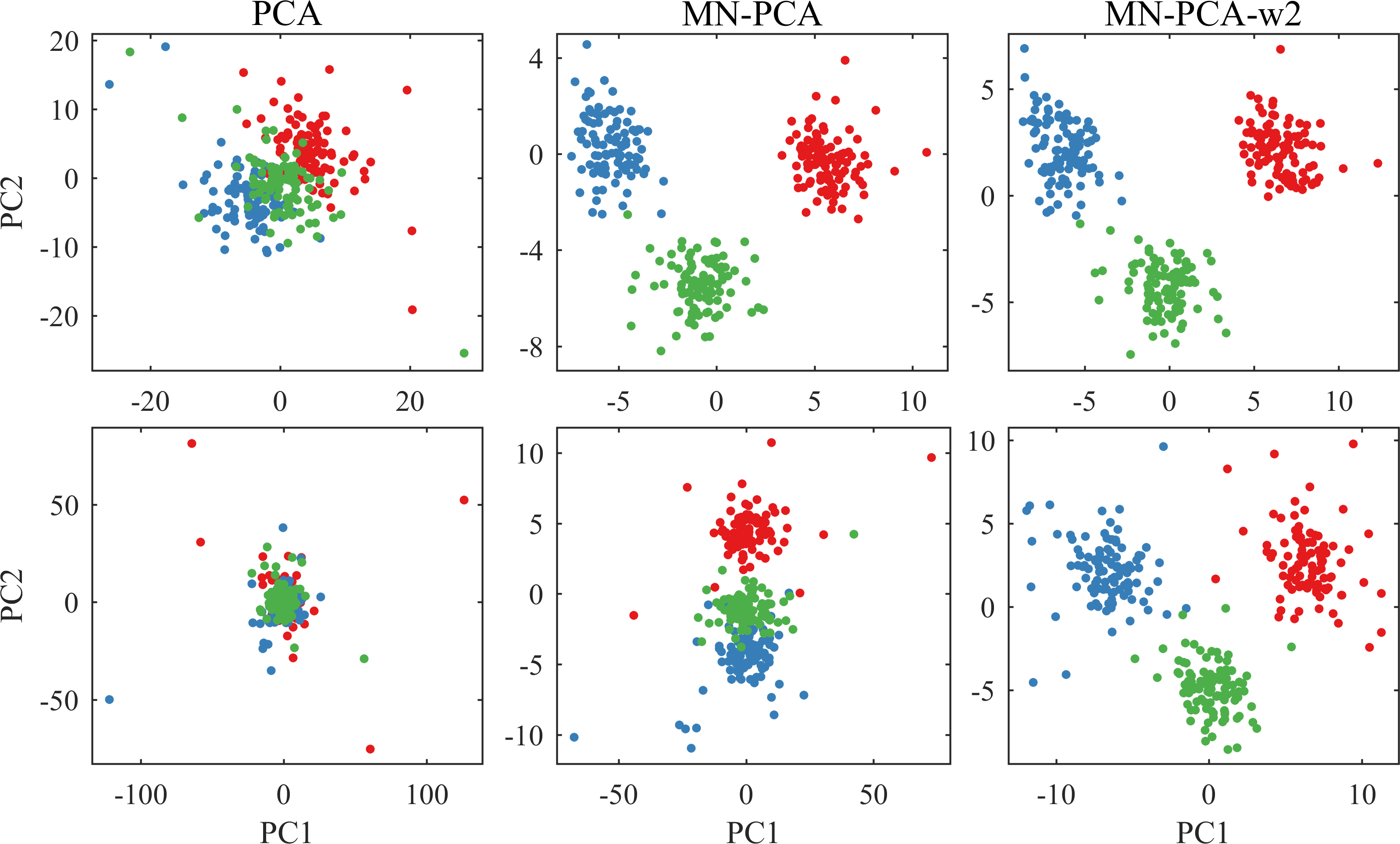}
    \caption{Illustration of PCA, MN-PCA and MN-PCA-w2 on small synthetic data. Top: $c=32$. Bottom: $c=224$.}
    \label{fig:syn_scatter}
\end{figure*}

 We illustrate the difference between MN-PCA and PCA on data generated with mild and large $c$, respectively (Fig. \ref{fig:syn_scatter}). We can see that both MN-PCA and MN-PCA-w2 have clear clusters when the condition number is mild ($c=32$). The bottom row shows that when the condition number is large ($c=224$), PCA has inferior performance. The projection of MN-PCA is more dispersed than that of MN-PCA-w2, suggesting that minimizing the Wasserstein distance is more accurate when the condition number is large.


\subsubsection{Large-Scale Synthetic Data}
\par To investigate the performance of the proposed methods on large-scale data, we construct a set of synthetic data with a similar strategy.
We generate data with $n=1000$, $k=2$, $c=196$, $\alpha_1=10^{-2}$, $\alpha_2=10^{-3}$ and vary $p$ from $2000$ to $6000$. The first 10\% features of centroid 1 equal one; the last 10\% features of centroid 2 equal one; the first 10\% features of centroid 3 equal one and the last 10\% features equal minus one.
The results are the average of 5 runs.

\par We show the performance of low-rank recovery on the large-scale synthetic data (Table \ref{table:large_lr}).
PCA fails to produce meaningful projection because the NMI is close to zero. MN-PCA-w2 significantly outperforms MN-PCA in terms of PSNR and RMSE. But MN-PCA has a better performance in clustering than that of MN-PCA-w2. It suggests that MN-PCA-w2 tends to recover the low-rank matrix more accurately, but the projection of MN-PCA is more dispersed and thus may have a better performance in clustering.

\par
\begin{figure}[!t]
    \centering
    \includegraphics[width=0.98\columnwidth]{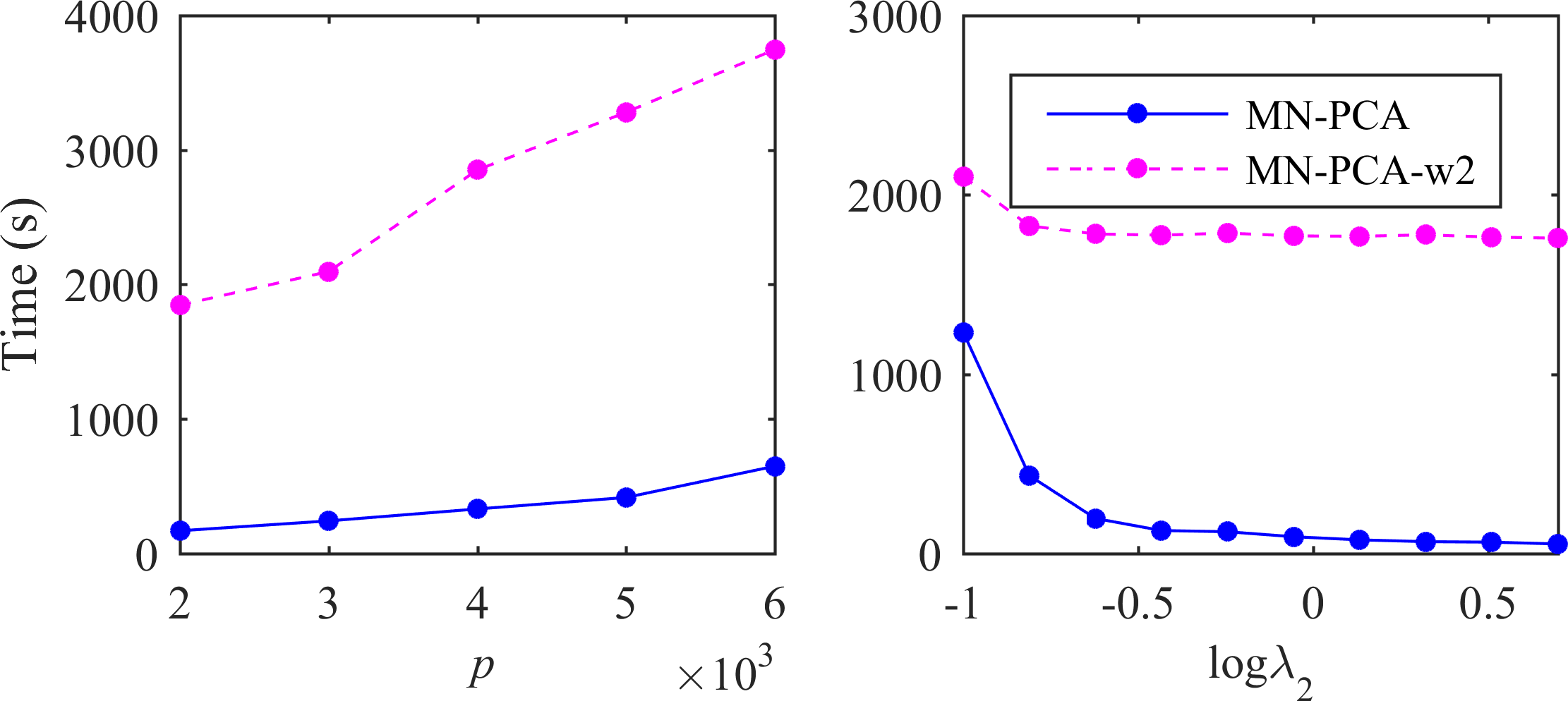}
    \caption{Running time of MN-PCA and MN-PCA-w2 on synthetic data of $n=1000$, $c=192$.
        Left: time versus data size.
        Right: time versus $\log \lambda_2$ with $p=2000$ and $\lambda_1=0.5$. }
    \label{fig:run_time}
\end{figure}

 We also show the running time of MN-PCA and MN-PCA-w2 (Fig. \ref{fig:run_time}).
MN-PCA is much faster than MN-PCA-w2 (left panel).
The running time of MN-PCA drops down rapidly with the increase of $\lambda_2$ (right panel).
Because a large $\lambda_2$ encourages the sparsity of the estimated precision matrix, allowing us to use the Cholesky decomposition on each block (see Sec \ref{sec:complexity}).
The running time of MN-PCA-w2 does not change a lot with the increase of $\lambda_2$, because its implementation does not fully exploit the sparsity of the estimated precision matrices.

\subsubsection{Swiss Roll Data}
\par
We then compare our methods with PCA on the Swiss roll data. Swiss roll data is widely used to test various dimensionality reduction algorithms. The idea of Swiss roll data is to create several points in 2D space and then map them to 3D space. Existing algorithms such as Isomap \cite{tenenbaum2000global}, LLE \cite{roweis2000nonlinear} and SDE \cite{weinberger2006unsupervised} considering the local geometry structure can unfold the roll to 2D space successfully.

\par
\begin{figure*}[!t]
    \centering
    \includegraphics[width=.85\textwidth]{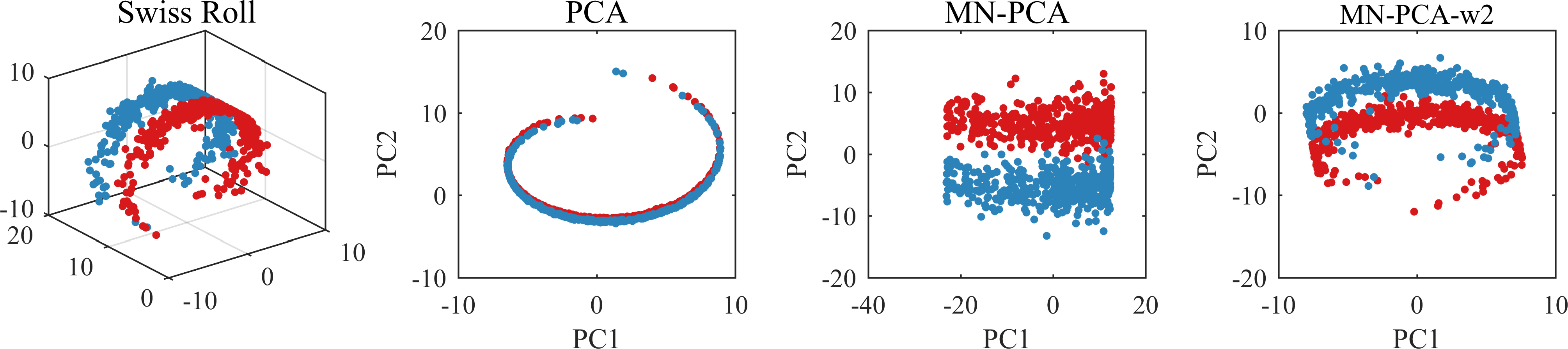}
    \caption{Illustration of PCA and MN-PCA on Swiss roll data. From left to right: the 3D scatter plot of Swiss roll data; the projection of PCA; the projection of MN-PCA; the projection of MN-PCA-w2}
    \label{fig:swiss_roll}
\end{figure*}

 We create two rolls, as shown in the first panel of Fig. \ref{fig:swiss_roll} and then use PCA and the proposed methods to project the data points to 2D space. Fig. \ref{fig:swiss_roll} presents that the two clusters in the projection of PCA are mixed.
But those two clusters are separated from the projection of the proposed methods.
MN-PCA does not consider the local geometry or geodesic distance explicitly. This experiment implies that the precision matrices somehow reflect the local structure and may guide us to obtain a better representation.

\subsection{Real-World Experiments}
\par In this section, we show the effectiveness of MN-PCA and compare it with PCA as well as three other competing methods on various real-world data, including kernel PCA (KPCA), ICA and GPCA.
\textbf{KPCA} is an extension of PCA, and it adopts the kernel idea to handle nonlinear data. However, how to choose an appropriate kernel for the given data is still unknown. Therefore, we apply three different kernels (i.e., linear, Gaussian, and polynomial) with recommended parameters to the data.
\textbf{ICA} has been well studied and extensively used in signal processing. We use an efficient and popular algorithm FastICA proposed by \cite{hyvarinen2000independent} for comparison. \textbf{GPCA} is the most competing approach \cite{Allen2014}.
However, it requires the precision matrix $\Omega^{-1}$ and $\Sigma^{-1}$ to be predefined. Two empirical ways have been suggested to set it: 1) the inverse smoothing matrix, which is computed by the Laplacian matrix $L$; 2) the standard exponential smoothing matrix $S$, which is based on the distance matrix. Finally, there are four different settings for GPCA, denoted as LL, LS, SL, and SS, respectively.
The statistics of the 18 used datasets  for evaluation are summarized in Table \ref{table:data_statistics}. More details are described in the Supplementary Materials.

\begin{table}[hpbt]
    \caption{Summary of the Datasets}
    \centering
    \begin{tabular}{lrrr}
        \toprule
        Datasets& \# Samples (n) & \# Features (p) & \# Classes \\
        \midrule
        balance-scale&  625& 4& 3 \\
        german&  1000& 20& 2 \\
        glass&  214& 9& 6 \\
        heart-statlog&  270& 13& 2 \\
        iono&  351& 34& 2 \\
        sonar&  208& 60& 2 \\
        tae&  151& 5& 3 \\
        vehicle&  846& 18& 4 \\
        wine&  178& 13& 3 \\
        wisc&  699& 9& 2 \\
        zoo&  101& 16& 7 \\
        METABRIC&  1975 & 1000& 5\\
        gse29505&  290 & 384 & 2\\
        3sources&  352 & 700 & 6\\
        gesture & 3003& 32 & 5\\
        eeg-eye & 5000 & 14 & 2\\
        mfeat & 2000 & 649 & 10\\
        radar & 3500 & 174& 7\\
        \bottomrule
    \end{tabular}\label{table:data_statistics}
\end{table}

\par All methods require an approximation of the rank of the data matrix. We adopt the bi-cross validation method \cite{owen2009bi}. 
Specifically, we first approximate the data matrix by the truncated SVD $\hat{Y_{k}}$, where $\hat{Y_{k}} = \sum_{j=1}^k \hat{d}_j \hat{u}_j \hat{v}_j^T$. The the proportion of variance explained is defined as $R_i = \sum_{j=1}^i \hat{d}_j^2 /\sum_{j=1}^k \hat{d}_j^2$.
Note that $R_i$ is between 0 and 1 and grows with $i$.
We assume that the redundant components should not contribute much to the total variance.
The final rank estimation is the smallest integer $k$, which satisfies $R_k > \tau$, $\tau>0$ is the thresholding proportion value (e.g. 0.8) and $1 \leq k \leq K$.

\par In the following experiments, we set $K=10$ and $\tau = 0.8$ to choose the rank for dimension reduction.
If the number of features are smaller than 15, we simply set $k=2$.
Then we use a Least-squares SVM (LSSVM) \cite{de2010ls} with a Gaussian kernel as a benchmark classifier to evaluate the quality of the low-rank presentation.
To facilitate a fair comparison, we choose the hyperparameter of LSSVM by 5-fold cross validation.
The accuracy is the average of 10 runs.
\begin{table*}[hpbt]
    \caption{Performance of Dimension Reduction on Real-world Datasets}
    \centering
    \begin{tabular}{lcccccccccc}
        \toprule
Dataset&PCA& GPLVM& KPCA (lin)& KPCA (gau)& ICA& GPCA (LS)& GPCA (SL)& MN-PCA& MN-PCA-w2\\
        \midrule
balance-scale& 66.37(0.76)& 68.05(0.74)& 52.51(0.74)& 63.34(0.34)& 61.94(0.63)& -& -& \textbf{77.39(1.11)}& \textbf{75.30(0.37)}\\
german& 71.02(0.65)& 70.31(0.36)& 71.05(0.74)& 70.25(0.14)& \textbf{72.71(0.60)}& 71.26(0.63)& -& 71.48(0.65)& \textbf{71.88(0.54)}\\
glass& 56.18(0.94)& \textbf{56.32(1.39)}& 51.54(2.03)& 32.14(1.91)& 16.96(1.01)& -& -& \textbf{61.01(2.18)}& 47.15(0.89)\\
heart-statlog& 80.33(0.95)& \textbf{82.96(0.65)}& \textbf{83.52(0.94)}& 74.11(1.79)& 76.56(0.25)& 74.56(1.16)& -& 82.07(0.66)& 82.70(0.72)\\
iono& 93.16(0.67)& \textbf{93.53(0.36)}& 92.99(0.57)& 92.02(0.43)& 89.86(0.61)& -& -& 91.17(0.43)& 92.09(0.67)\\
sonar& \textbf{85.83(2.21)}& 71.16(2.49)& \textbf{85.54(2.34)}& 59.09(3.35)& 49.66(2.05)& 85.44(1.32)& -& 85.34(1.68)& 81.15(2.15)\\
tae& 43.13(4.47)& 42.71(5.00)& 40.71(2.77)& \textbf{54.83(3.37)}& 46.49(3.22)& -& -& \textbf{49.55(3.11)}& 42.43(4.35)\\
vehicle& 45.72(0.64)& 46.60(0.71)& 43.83(0.89)& 28.90(0.74)& 9.79(0.58)& 50.07(1.38)& -& \textbf{51.72(0.93)}& \textbf{58.17(0.93)}\\
wine&\textbf{ 95.95(0.24)}& 95.79(0.60)& 95.79(0.40)& 90.93(1.25)& 56.29(2.01)& 63.82(1.72)& -& \textbf{96.74(0.44)}& 93.09(0.28)\\
wisc& \textbf{96.85(0.13)}& 96.81(0.14)& \textbf{96.87(0.17)}& 95.95(0.14)& 91.39(0.55)& -& -& 96.50(0.24)& 96.15(0.17)\\
zoo& \textbf{89.51(1.25)}& 85.72(1.38)& 87.50(1.95)& 66.77(2.66)& 73.77(3.26)& -& 88.63(2.10)& 76.45(4.61)& \textbf{90.46(5.21)}\\
metabric& 66.12(0.28)& 65.85(0.32)& \textbf{67.08(0.32)}& 4.52(0.21)& 1.58(0.24)& 50.96(0.86)& 62.49(0.46)& \textbf{67.74(0.62)}& 61.25(0.34)\\
gse29505& 92.86(0.56)& 92.86(0.56)& 92.59(0.55)& 59.90(0.54)& 48.97(3.25)& 87.07(0.97)& \textbf{95.76(0.73)}& \textbf{93.24(0.52)}& 92.31(0.61)\\
3sources& 70.55(1.82)& 69.89(0.99)& \textbf{70.59(1.22)}& 2.47(0.46)& 0.03(0.09)& -& 14.26(1.31)& \textbf{76.78(0.88)}& 63.15(1.26)\\
gesture & 54.49(0.16) & 54.96(0.47) & 54.82(0.32) & 53.25(0.31) & 51.75(0.34) & \textbf{56.87(0.45)} & - & 55.42(0.32)&\textbf{57.72(0.46)}
 \\
eeg-eye & 61.94(0.27) & 62.22(0.30) & 65.93(0.18) & \textbf{66.17(0.18)} & \textbf{66.10(0.23)} & - & - & 64.25(0.44)&  65.54(0.30) \\
mfeat    & 90.53(0.28) & \textbf{91.55(0.21)} & 88.71(0.35) & 0.55(0.14)  & 1.59(0.29)  & - & - & \textbf{90.71(0.41)}& 69.54(1.04)\\
radar & 95.57(0.23) & \textbf{96.17(0.20)} & 95.47(0.19) & 7.51(0.18) & 4.09(0.19) & - &- & \textbf{95.84(0.13)} & 89.80(0.32)\\
        \bottomrule
    \end{tabular}\label{table:real_world}
        \begin{tablenotes}
    \small
    \item Top two of the cross-validated accuracies are shown in bold. GPCA fails to produce results when the predefined precision matrices are ill-conditioned. Due to space limitations, the results of PPCA, KPCA with polynomial kernel, GPCA(LL) and GPCA(SS)   are shown in the Supplementary Materials.
\end{tablenotes}
\end{table*}

\par
We demonstrate the cross-validated classification accuracies of these methods on the real-world datasets (Table \ref{table:real_world}). 
The running time is reported in Supplementary Materials.
We could clearly find the following observations.
1) Our methods have competitive or superior performance compared with PCA in most cases, suggesting that MN-PCA could obtain more informative representation in some real applications.
2) The performance of GPCA is poorer than MN-PCA in general. Although GPCA with the predefined precision matrices was considered to work well on the brain MRI data, it needs to choose appropriate precision matrices in advance, which is not suitable for general data.
3) The performance of GPLVM is similar to PCA.
4) The performance of KPCA with different kernels are very different.
5) Both KPCA and GPLVM adopt the kernel trick to handle the nonlinear data, but it is challenging to choose an appropriate kernel for a given data.
6) ICA sometimes produces inferior performance. A possible reason is that the non-Gaussian assumption of ICA is too restrictive.

\subsection{The Estimated Precision Matrix is Informative}
\begin{figure}[!htp]
    \centering
    \includegraphics[width=.74\columnwidth]{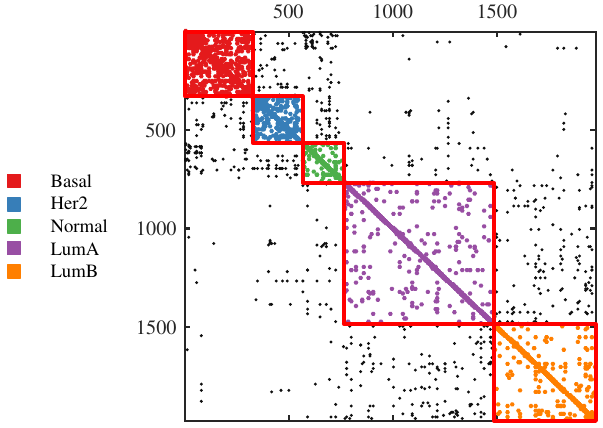}
    \caption{Illustration of the estimated precision matrix of MN-PCA in the sample space. It is reordered based on the subtypes with $k = 5$. Since the estimated precision matrix contains many small values which should be less significant, only top 1000 interactions are shown.}
    \label{fig:metabric_iA}
\end{figure}

\par We have shown the effectiveness of MN-PCA in terms of dimension reduction and low-rank representation.
Furthermore, the estimated precision matrices also reveal some inspiring patterns in the noise.
Take the METABRIC data as an illustrating example.
It contains a large cohort of around 2000 breast cancer patients with detailed clinical measurements and genome-wide molecular profiles.
How to use the gene expression profile to classify invasive breast cancer into biologically and clinically distinct subtypes has been intensively studied.
Here we use the famous PAM50 subtypes as the reference \cite{parker2009supervised} to evaluate the low-rank representation and graphical noise structure.
We select the top 1000 genes by the coefficients of variation and focus on the precision matrix in the sample space.
There are 5504 edges in the estimated precision matrix (Fig \ref{fig:metabric_iA}).
It demonstrates that the interactions within subtypes tend to be denser than that between subtypes, and biologically relevant subtypes have more interactions. For example, LumA and LumB have more interactions. Normal-like subtypes have many interactions with the other subtypes.

\par The next example is about 3sources data, which is collected from three well-known online news sources, including BBC, Reuters, and The Guardian. Here we use the documents collected from the BBC as an example. The document term matrix is of size $352 \times 3560$. We remove the terms that appear less than 20 times in documents. There are 700 terms left. We focus on the estimated precision matrix in the feature space, which reveals the relationship of terms in the documents. The estimated rank $k=7$, and there are 651 edges. We note that the semantically related terms tend to have interactions. For example, the top 5 term (chosen by absolute values) pairs are "premier/league", "study/research",  "executive/chief", "minister/primer", "journal/study", respectively. Those term pairs frequently appear in the same documents.

\section{Discussion and Conclusion}
\par
We propose MN-PCA to model the graphical noise in both the feature and sample spaces. MN-PCA can be regarded as the minimization of the reconstruction error over the generalized Mahalaobis distance.
PCA is a special case of our model.
We develop two algorithms for inference.
The first one is to maximize the regularized likelihood that works well on the synthetic data when the effect of the matrix normal noise is mild and can handle relatively large data.
But it is not robust when the condition numbers get larger.
To address this challenge, we propose to minimize the Wasserstein distance between the the transformed residue and the white noise.
It is more robust when the effect of the matrix normal noise is substantial.
Extensive numerical results show that considering the structural noise brings an improvement in classification,  suggesting our methods find better low-rank representations.
Moreover, the inferred precision matrices are informative and can help us understand the underlying structure of the noise.

\par  There are several questions that remain to be investigated.
First, the estimation of the precision matrices is computationally expensive, which makes the proposed algorithms inefficient for large data.
It is worth studying how to develop a more efficient and robust algorithm for big data.
Second, our proposed methods are not guaranteed to provide a better low-rank representation for clustering and classification.
As we demonstrate in the real-world experiments,  the estimated precision matrices are also informative.
If the inferred structure of the noise captures the information related to clustering and classification, the performance of the low-rank representation possibly decreases.
 One crucial problem is that what is the noise, and what is the signal?
 Conventional methods such as PCA assume the noise is white noise, which is seldom satisfied.
 In this paper, we model the graphical noise by matrix normal distribution. Although it is difficult to determine what information is captured in the estimated noise without external knowledge, our methods provide an approach for data exploration for users to obtain the low-rank representation and discover the underlying structure of the noise simultaneously.

\par MN-PCA can be extended from two aspects.
First, it can be extended to matrix variate data, where one has multiple matrix variate observations.
Second, it is worth to generalize MN-PCA for tensor data by the tensor normal distribution, enabling us to explore the correlation of the noise in the high-order data.


%


\ifCLASSOPTIONcaptionsoff
  \newpage
\fi



%
\bibliographystyle{IEEEtran}
\bibliography{IEEEabrv,reference}

%





\end{document}


\maketitle
\tableofcontents

\newpage
\section{Methods}
\subsection{Proof of Theorem 1}
\begin{proof}
	The expectation of the sample covariance is as follows:
	\begin{equation}
	\mathbb{E}\left(\frac{1}{n}Y^TY\right) = \frac{1}{n} M^TM + \frac{1}{n}\tr(\Omega)\Sigma.
	\end{equation}
	By the eigenvalue stability inequality \cite{Tao2012}, we have the following upper bound:
	\begin{equation}\label{eq:pca_noise_effect}
\left|\sigma_i\left(\frac{1}{n}\mathbb{E}(Y^TY)\right) - \sigma_i\left(\frac{1}{n}M^TM\right)\right|\leq \norm{\frac{1}{n}\tr(\Omega)\Sigma}_{op}=\sigma\sigma_1(\Sigma),
\end{equation}
where $\norm{\cdot}_{op}$ is the operator norm of matrix.
The last equation holds because the operator norm of matrix $\Omega$ can be computed by the square root of the largest eigenvalue of $\Omega^T\Omega$.

\end{proof}

\subsection{Proof of Theorem 2}
\begin{proof}
	Since the number of rows of $QER$ equals to $|S|$, the optimal transport map from the rows of $QER$ to $S$ always exists.
	Let $T$ denote the optimal transport map from the rows of $QER$ to $S$, then by the definition of the Wasserstein distance,
	\begin{equation}
	W^2(\mbox{row}\{QER\},S) = \frac{1}{n}\norm{QER - T(QER)}_F^2,
	\end{equation}
	where $T(QER)$ is a $n\times p$ matrix with its rows corresponding to those of $QER$ by $T$. Actually, we can transform $T(QER)$ to $H$ by rearranging its rows, i.e., $T(QER)=PH$, where $P$ is a permutation matrix. Then
	\begin{align}
	\frac{1}{n}\norm{QER - T(QER)}_F^2 &= \frac{1}{n} \norm{P^TQER - P^T T(QER)}_F^2\\
	& = \frac{1}{n}\norm{P^TQER - H}_F^2.
	\end{align}
	Since $M$ is obtained by SVD of $Y$ and $E=Y-M$, the rank of E is $p-k$. 
	Hence, $\frac{1}{n}\norm{P^TQER - H}_F^2$ is minimized if and only if:
	\begin{equation}
	P^TQER = H_{p-k},
	\end{equation}
	which completes the proof.
\end{proof}

\subsection{GPCA Post-processing}
As aforementioned in the main manuscript, the solution of $X$ and $W$ is not unique. Specifically, given any invertible matrix $P$, $XPP^{-1}W^T$ equals $XW^T$. Therefore, $X$ and $W$ are not identifiable. To address this issue, we use GPCA \cite{allen2014generalized} to post-process $W$ and $X$, and obtain the unique solutions of $X$ and $W$. The objective function of GPCA is as follows:
\begin{align}
 \begin{split}\label{eq:gpca}
 \min \ & \frac{1}{2} \tr\left(Y - UDV^T\right)\Omega^{-1} \left(Y - UDV^T\right)^T \Sigma ^{-1}\\
 \text{s.t. }    & U^T\Sigma ^{-1}U = I\\
 & V^T\Omega^{-1}V = I \\
 & \text{diag}(D) \geq 0.
 \end{split}
\end{align}

\par Recall that PCA can be interpreted as the minimization of the reconstruction error under the Frobenius norm. Given $Y$, PCA aims at finding a low-rank approximation such that
\begin{equation}
   \min \norm{Y - XW^T}_F^2.
\end{equation}
Note that the objection function of GPCA can be rewritten as
\begin{equation}
 \norm{\Omega^{-\frac{1}{2}}(Y - UDV^T)\Sigma ^{-\frac{1}{2}}}_F^2,
\end{equation}
when $\Omega^{-1}$ and $\Sigma ^{-1}$ are positive definite, $\norm{\Omega^{-\frac{1}{2}} X \Sigma ^{-\frac{1}{2}} }_F$ is a proper matrix norm for any $X\in R^{n \times p}$. Let's denote this generalized matrix norm as $\Omega^{-1}, \Sigma ^{-1}$-norm, or $\norm{X}_{\Omega^{-1}, \Sigma ^{-1}}$. For simplicity, we also denote $\norm{X}_{\Omega^{-1}, I}$ as $\norm{X}_{\Omega^{-1}}$ and $\norm{X}_{I, \Sigma ^{-1}}$ as $\norm{X}_{\Sigma ^{-1}}$, where $I$ is the identity matrix of a proper size.

\par Therefore, GPCA can be interpreted as the minimization of the reconstruction error under the $\Omega^{-1}, \Sigma ^{-1}$-norm. Inspired by PCA, GPCA employs a generalized power method to obtain the columns of $U$ and $W$ sequentially. The scheme is given in \textbf{Algorithm \ref{alg:post}}.
\begin{table}[hpbt]
 \begin{algorithm}[H]
    \begin{algorithmic}[1]\caption{\textbf{GPCA (Post-processing)}}\label{alg:post}
        \Input  data matrix $Y$, precision matrices $\Omega^{-1}$, $\Sigma ^{-1}$ and initialization of $u_1$ and $v_1$
        \Output  $U$, $D$, $V$
        \State $Y^{(1)} = X$
        \For {$r=1,\cdots, k$}
        \Repeat
        \State set $u_k = \frac{ Y^{(r)}\Sigma ^{-1}v_k }{ \norm{ Y^{(r)}\Sigma ^{-1}v_k }_{\Omega^{-1} }}$
        \State set $v_k = \frac{ Y^{(r)T}\Omega^{-1}u_k }{ \norm{ Y^{(r)T}\Omega^{-1}u_k }_{\Sigma ^{-1} }}$
        \Until{convergence}
        \State set $d_k = u_k^T \Omega^{-1} Y^{(r)} \Sigma ^{-1}v_k$
        \State set $Y^{(r+1)} = Y^{(r)} - u_k d_k v_k^T$
        \EndFor
        \State set $U = [u_1, \dots, u_k]$, $V = [v_1, \cdots, v_k]$ and $D = \text{diag}(d_1, \dots, d_k)$
    \end{algorithmic}
\end{algorithm}
\end{table}

\subsection{Minimizing Wasserstein Distance and its Relaxation}
\par The covariance of $\mbox{vec}(Y-M)$ is $\Omega\otimes\Sigma$. Taking it into Eq. (42) in the main text and adding $L_1$-norm regularizers, the objective function of minimizing Wasserstein distance should be:
\begin{equation}\label{eq:w2_obj}
\min_{Q,R}\frac{1}{np}\|QER\|_{2}^{4}-\frac{2\sigma}{\sqrt{np}}\|QER\|_{*}^{2}+ \lambda_{1}\|Q^{T}Q\|_{1}+ \lambda_{2}\|R^{T}R\|_{1}.
\end{equation}
The objective function is not homogeneous on $\sigma$. Therefore, the original function is sensitive to the hyperparameter $\sigma$. But it is difficult to choose or estimate an appropriate value of $\sigma$ for a given data. To overcome this limitation, we relax the original objective function by applying the square root of the first two terms:
\begin{equation}\label{eq:w2_obj_relax}
    \min_{Q,R}\frac{1}{\sqrt{np}}\|QER\|_{2}^{2}-\frac{2}{\sqrt[4]{np}}\|QER\|_{*}+ \lambda_{1}\|Q^{T}Q\|_{1}+ \lambda_{2}\|R^{T}R\|_{1}.
\end{equation}

\par Now the objective function is homogeneous on $\sigma$. Specifically, we can see that
\begin{equation}
Q^{*},R^{*}=\mathop{\arg\min}_{Q,R}\frac{1}{\sqrt{np}}\|QER\|_{2}^{2}-\frac{2}{\sqrt[4]{np}}\|QER\|_{*},
\end{equation}
then
\begin{equation}
\sqrt{\sigma}Q^{*},\sqrt{\sigma}R^{*}=\mathop{\arg\min}_{Q,R}\frac{1}{\sqrt{np}}\|QER\|_{2}^{2}-\frac{2\sigma}{\sqrt[4]{np}}\|QER\|_{*}.
\end{equation}
In addition, the first term of Eq. (\ref{eq:w2_obj}) is quartic to $Q$ and $R$, which is difficult to solve. The first term of Eq. (\ref{eq:w2_obj_relax}) becomes quardratic, which can be solved more efficiently. For clarity, we refer the original version as MN-PCA-w2*. We implement both the algorithms based on the gradient decent strategy with PyTorch \cite{paszke2017automatic}. We add the experimental results of MN-PCA-w2* and its relaxation on both synthetic data and real-world data in Section \ref{sec:exp_res}.

\subsection{Performance Metrics}
The true positive rate, true negative rate and predictive positive value are defined as follows:
\begin{equation}
    \text{TPR} = \frac{\#\{ \hat{\Omega}_{ij} \neq 0 \ \& \ \Omega_{ij} \neq 0\} }{\#\{ \Omega_{ij} \neq 0\} },
\end{equation}
\begin{equation}
\text{TNR} = \frac{\#\{ \hat{\Omega}_{ij} = 0 \ \& \ \Omega_{ij} = 0\} }{\#\{ \Omega_{ij} = 0\}},
\end{equation}
\begin{equation}
\text{PPV} = \frac{\#\{ \hat{\Omega}_{ij} \neq 0 \ \& \ \Omega_{ij} \neq 0\} }{\#\{ \hat{\Omega}_{ij} \neq 0\} }.
\end{equation}

\section{Experimental Results}\label{sec:exp_res}
\subsection{Comparison on Synthetic Data}
\par We apply MN-PCA-w2* to the small synthetic datasets. The generation process is described in the main manuscript. The comparison results of MN-PCA-w2* and MN-PCA-w2 are shown in Table \ref{table:syn_small_recovery}. We can find that
\begin{itemize}
    \item When $c=8$ and $c=16$, MN-PCA-w2* has significantly poorer performance and very large standard deviation compared to MN-PCA. This is due to that MN-PCA-w2* raises errors when computing SVD.
          When the condition number $c$ is small, the residue is relatively small in the Frobenius norm.
          It makes the computation of SVD unstable.
    \item MN-PCA-w2 tends to outperform MN-PCA-w2*, and its performance has smaller standard deviation.
    Note that the underlying $\sigma$ is known, so we can simply set the $\sigma$ of MN-PCA-w2* as the ground truth.
    This may contribute to the robustness of the performances of MN-PCA-w2*.
    However, the true value of $\sigma$ is unknown when we apply MN-PCA-w2* to the real-world data.
    We find that estimating the value of $\sigma$ approximately is still difficult.
\end{itemize}

\begin{table}[hpbt]
	\centering
	{\resizebox{\textwidth}{!}{\begin{minipage}{\textwidth}
				\begin{threeparttable}
					\caption{Comparison of the low-rank recovery on the small synthetic datasets}
					\begin{tabular}{lrrrrrr}
						\toprule
						& \multicolumn{2}{c}{PSNR} & \multicolumn{2}{c}{RMSE} &  \multicolumn{2}{c}{NMI}\\
						\cmidrule(lr){2-3} \cmidrule(lr){4-5} \cmidrule(lr){6-7}
						&MN-PCA-w2* &MN-PCA-w2  &MN-PCA-w2* &MN-PCA-w2 &MN-PCA-w2*& MN-PCA-w2 \\
						\midrule
						$c=8$   & 9.50(3.11)  & \textbf{16.64(0.17)} & 0.36(0.15)          & \textbf{0.15(0.00)} & 52.99(30.48) & \textbf{98.22(1.69)}  \\
						$c=16$  & 9.77(2.99)  & \textbf{16.28(0.23)} & 0.35(0.14)          & \textbf{0.15(0.00)} & 58.26(31.17) & \textbf{99.30(0.90)}  \\
						$c=32$  & 11.56(1.06) & \textbf{13.12(3.55)} & 0.27(0.03)          & \textbf{0.24(0.10)} & 78.57(11.56) & \textbf{82.88(18.82)} \\
						$c=64$  & 10.00(1.03) & \textbf{11.36(3.34)} & 0.32(0.04)          & \textbf{0.29(0.10)} & 67.31(10.75) & \textbf{75.40(19.51)} \\
						$c=96$  & 9.32(0.67)  & \textbf{10.37(3.02)} & 0.34(0.03)          & \textbf{0.32(0.09)} & 61.99(5.11)  & \textbf{70.16(18.95)} \\
						$c=128$ & 9.31(1.23)  & \textbf{9.58(3.40)}  & \textbf{0.35(0.05)} & 0.35(0.12)          & 63.65(14.77) & \textbf{58.78(34.88)} \\
						$c=160$ & 8.50(1.33)  & \textbf{9.33(3.19)}  & 0.38(0.06)          & \textbf{0.36(0.12)} & 56.41(19.32) & \textbf{59.25(34.74)} \\
						$c=192$ & 8.46(0.81)  & \textbf{9.62(3.24)}  & 0.38(0.03)          & \textbf{0.35(0.14)} & 62.34(9.88)  & \textbf{68.71(28.93)} \\
						$c=224$ & 7.97(0.83)  & \textbf{8.99(3.28)}  & 0.40(0.04)          & \textbf{0.38(0.14)} & 59.03(8.04)  & \textbf{63.86(27.14)}\\
						\bottomrule
					\end{tabular}\label{table:syn_small_recovery}
				\end{threeparttable}
				\begin{tablenotes}
					\small
					\item The best performance is shown in bold.
				\end{tablenotes}
		\end{minipage}}
	}
\end{table}

\par We compare the performance of estimating the precision matrices in Table \ref{table:syn_small_graph}. The definition of the used metrics are in the main manuscript. We can see that MN-PCA-w2 achieves better performance than MN-PCA-w2* in terms of all metrics.

\par In summary, the relaxation of MN-PCA-w2* is homogeneous on $\sigma$ and thus eliminate the unnecessary hyperparameter. Moreover, the first term of the objective function of MN-PCA-w2* is quartic to $Q$ and $R$, which can be difficult. MN-PCA-w2 applies the square root to the first two terms of the objective function of MN-PCA-w2*. The quartic term becomes quadratic, which can be solved more efficiently. The relaxation of MN-PCA-w2* has better performance on the small synthetic data.

\begin{table}[hpbt]
    \centering
    \begin{adjustbox}{width=1\textwidth}
                \begin{threeparttable}
                    \caption{Comparison of the precision matrix estimation on the small synthetic datasets}
                    \label{table_syn_lowrank}
                    \begin{tabular}{lrrrrrr}
                        \toprule
                        & \multicolumn{2}{c}{$\text{TPR}_1$} & \multicolumn{2}{c}{$\text{TNR}_1$} & \multicolumn{2}{c}{$\text{PPV}_1$}\\
                        \cmidrule(lr){2-3} \cmidrule(lr){4-5} \cmidrule(lr){6-7}
                        &MN-PCA-w2* &MN-PCA-w2  &MN-PCA-w2* &MN-PCA-w2 &MN-PCA-w2*& MN-PCA-w2 \\
                        \midrule
                        $c=8$   & 9.50(3.11)  & \textbf{16.64(0.17)} & 0.36(0.15)          & \textbf{0.15(0.00)} & 52.99(30.48) & \textbf{98.22(1.69)}  \\
                        $c=16$  & 9.77(2.99)  & \textbf{16.28(0.23)} & 0.35(0.14)          & \textbf{0.15(0.00)} & 58.26(31.17) & \textbf{99.30(0.90)}  \\
                        $c=32$  & 11.56(1.06) & \textbf{13.12(3.55)} & 0.27(0.03)          & \textbf{0.24(0.10)} & 78.57(11.56) & \textbf{82.88(18.82)} \\
                        $c=64$  & 10.00(1.03) & \textbf{11.36(3.34)} & 0.32(0.04)          & \textbf{0.29(0.10)} & 67.31(10.75) & \textbf{75.40(19.51)} \\
                        $c=96$  & 9.32(0.67)  & \textbf{10.37(3.02)} & 0.34(0.03)          & \textbf{0.32(0.09)} & 61.99(5.11)  & \textbf{70.16(18.95)} \\
                        $c=128$ & 9.31(1.23)  & \textbf{9.58(3.40)}  & \textbf{0.35(0.05)} & 0.35(0.12)          & \textbf{63.65(14.77)} & 58.78(34.88) \\
                        $c=160$ & 8.50(1.33)  & \textbf{9.33(3.19)}  & 0.38(0.06)          & \textbf{0.36(0.12)} & 56.41(19.32) & \textbf{59.25(34.74)} \\
                        $c=192$ & 8.46(0.81)  & \textbf{9.62(3.24)}  & 0.38(0.03)          & \textbf{0.35(0.14)} & 62.34(9.88)  & \textbf{68.71(28.93)} \\
                        $c=224$ & 7.97(0.83)  & \textbf{8.99(3.28)}  & 0.40(0.04)          & \textbf{0.38(0.14)} & 59.03(8.04)  & \textbf{63.86(27.14)}\\
              \midrule
               & \multicolumn{2}{c}{$\text{TPR}_2$} & \multicolumn{2}{c}{$\text{TNR}_2$} & \multicolumn{2}{c}{$\text{PPV}_2$}\\
                \cmidrule(lr){2-3} \cmidrule(lr){4-5} \cmidrule(lr){6-7}
$c=8$   & 0.09(0.05) & \textbf{0.17(0.03)} & 0.99(0.00) & 0.99(0.00) & 0.12(0.07) & \textbf{0.20(0.02)} \\
$c=16$  & 0.13(0.07) & \textbf{0.21(0.05)} & 0.99(0.00) & 0.99(0.00) & 0.16(0.08) & \textbf{0.24(0.04)} \\
$c=32$  & 0.19(0.04) & \textbf{0.24(0.05)} & 0.99(0.00) & 0.99(0.00) & 0.22(0.04) & \textbf{0.26(0.03)} \\
$c=64$  & 0.20(0.04) & \textbf{0.26(0.06)} & 0.99(0.00) & 0.99(0.00) & 0.23(0.03) & \textbf{0.28(0.04)} \\
$c=96$  & 0.21(0.04) & \textbf{0.28(0.08)} & 0.99(0.00) & 0.99(0.00) & 0.24(0.04) & \textbf{0.29(0.05)} \\
$c=128$ & 0.20(0.05) & \textbf{0.28(0.07)} & 0.99(0.00) & 0.99(0.00) & 0.23(0.04) & \textbf{0.28(0.05)} \\
$c=160$ & 0.21(0.04) & \textbf{0.28(0.08)} & 0.99(0.00) & 0.99(0.00) & 0.24(0.04) & \textbf{0.28(0.05)} \\
$c=192$ & 0.20(0.04) & \textbf{0.30(0.07)} & 0.99(0.00) & 0.99(0.00) & 0.23(0.04) & \textbf{0.30(0.05)} \\
$c=224$ & 0.20(0.04) & \textbf{0.30(0.07)} & 0.99(0.00) & 0.99(0.00) & 0.23(0.04) & \textbf{0.30(0.05)}\\
                 \bottomrule
                    \end{tabular}\label{table:syn_small_graph}
                \end{threeparttable}
\end{adjustbox}
\end{table}

\subsection{Real-world Datasets}
\par We used 18 real-world datasets in the experiments. 15 of them were downloaded from the UCI machine learning repository, including:

\par \textbf{balance scale\footnote{http://archive.ics.uci.edu/ml/datasets/balance+scale.}:}
This data set was generated to model the psychological experimental results.

\par \textbf{german\footnote{https://archive.ics.uci.edu/ml/datasets/statlog+(german+credit+data)}:}
This dataset classifies people described by a set of attributes as good or bad credit risks.

\par
\textbf{glass\footnote{https://archive.ics.uci.edu/ml/datasets/glass+identification}:}
This dataset contains 6 types of glasses defined in terms of their oxide content (i.e. Na, Fe, K, etc).

\par
\textbf{heart-statlog\footnote{http://archive.ics.uci.edu/ml/datasets/statlog+(heart)}:}
This dataset contains the information of patients to predict the heart disease.

\par
\textbf{iono\footnote{https://archive.ics.uci.edu/ml/datasets/ionosphere}:}
Ionosphere dataset was collected from the radar returns. It contains two types of samples: ``Good" radar returns that shows the evidence of some type of structure in the ionosphere. ``Bad" returns are that do not.

\par
\textbf{sonar\footnote{http://archive.ics.uci.edu/ml/datasets/connectionist+bench+(sonar,+mines+vs.+rocks)}:}
This dataset contains sonar signals that bounced off a metal signals and that bounced off a roughly cylindrical rock.

\par
\textbf{tae\footnote{https://archive.ics.uci.edu/ml/datasets/teaching+assistant+evaluation}:}
Teaching assistant evaluation dataset consists of the evaluations of teaching assistant assignments.

\par
\textbf{vehicle\footnote{https://archive.ics.uci.edu/ml/datasets/Statlog+(Vehicle+Silhouettes)}:}
Vehicle silhouettes dataset contains the features extracted from the silhouettes of vehicle images. The aim of this data is to classify a given silhouette as one of four types of vehicles.

\par
\textbf{wine\footnote{https://archive.ics.uci.edu/ml/datasets/wine}:}
This dataset is to use chemical analysis to determine the origin of wines.

\par
\textbf{wisc\footnote{https://archive.ics.uci.edu/ml/datasets/breast+cancer+wisconsin+(original)}:}
The breast cancer Wisconsin (original) dataset consists of the clinical indicators of breast cancer samples and the corresponding classes.

\par
\textbf{zoo\footnote{https://archive.ics.uci.edu/ml/datasets/zoo}:}
The zoo dataset consists of the attributes of different kinds of animals and the classes.

\par
\textbf{gesture\footnote{https://archive.ics.uci.edu/ml/datasets/gesture+phase+segmentation}:} The dataset is composed by features extracted from 7 videos with people gesticulating, aiming at studying Gesture Phase Segmentation. We used a subset of this data.

\textbf{eeg-eye\footnote{https://archive.ics.uci.edu/ml/datasets/EEG+Eye+State}:} This dataset is from one continuous EEG measurement with the Emotiv EEG Neuroheadset. The duration of the measurement was 117 seconds. The eye state was detected via a camera during the EEG measurement. We used a subset of this data.
\par
\textbf{mfeat\footnote{https://archive.ics.uci.edu/ml/datasets/Multiple+Features}:} This dataset consists of features of handwritten numerals (`0'--`9') extracted from a collection of Dutch utility maps. 200 patterns per class (for a total of 2,000 patterns) have been digitized in binary images. The features are extracted by different algorithms and there are  649 features in total.
\par
\textbf{mfeat\footnote{https://archive.ics.uci.edu/ml/datasets/Multiple+Features}:} This dataset consists of features of handwritten numerals (`0'--`9') extracted from a collection of Dutch utility maps. 200 patterns per class (for a total of 2,000 patterns) have been digitized in binary images. The features are extracted by various algorithms. There are  649 features in total.
\par
\textbf{radar\footnote{https://archive.ics.uci.edu/ml/datasets/Crop+mapping+using+fused+optical-radar+data+set}:} This data set is a fused bi-temporal optical-radar data for cropland classification. It contains 7 classes. We sample 500 instances for each class to test our algorithm.

\par We used the same procedure to process those datasets. Specifically, we first removed the features that are not numerical. Then we removed the samples containing missing values. Finally, we normalized all the features by the z-score strategy to avoid the problem that different features may have different scales. The remaining 3 datasets are:

\par
\textbf{gse26505\footnote{https://www.ncbi.nlm.nih.gov/geo/}:}
This dataset is available with accession (GSE26505). It consists of 290 samples (colon, breast, lung, thyroid and Wilims' tumor cancers) and matched normal samples. Each sample has 384 methylation probes. We used the same procedure to process this data as that in \cite{chen2013biclustering}.

\par The details and the preprocessing procedure of the \textbf{METABRIC} and \textbf{3sources} are described in the main manuscript Section 4.3. After preprocessing the data, we used a standard procedure to choose the rank $k$. The chosen $k$ of each dataset is shown in Table \ref{table:chosen_r}.
\begin{table}[hpbt]
    \centering
    \caption{Selection of $k$ on the real-world datasets}
    \label{table:chosen_r}
    \begin{tabular}{lc|lc}
        \toprule
        Dataset & $k$ & Dataset & $k$ \\
        \midrule
        balance-scale & 2 & vehicle & 3 \\
        german & 8 & wine & 2\\
        glass & 2 & wisc & 2 \\
        heart-statlog& 2 & zoo & 5 \\
        iono & 6 & METABRIC & 5\\
        sonar & 6 & gse29505 & 5\\
        tae& 2& 3sources &  7\\
gesture & 7 & mfeat & 6 \\
eeg-eye & 2 & radar & 5\\
        \bottomrule
    \end{tabular}
\end{table}

\par The additional results of methods those are not contained in the main manuscript are reported in Table \ref{table:real-world}, including PPCA, KPCA(pol), GPCA(SS) and GPCA(LL).
In general, the performance of PPCA is similar to that of PCA.
GPCA(SS) and GPCA(LL) tend to fail because the predefined precision matrices are ill-conditioned.
The performance of KPCA(pol) is also not robust. It may produce poor performance on some data. For example on 3sources data, the performance of KPCA(pol) is significantly poorer than that of PCA. The performance of the original version of MN-PCA-w2* is also not very robust compared to the relaxed version MN-PCA-w2.
\begin{table}[hpt]
    \centering
    \caption{Additional results on the real-world datasets}
    \label{table:real-world}
    \begin{tabular}{lcccc}
     \toprule
     Dataset &PPCA& KPCA(pol) & GPCA(LL) & GPCA(SS) \\
     \midrule
balance-scale&58.93(0.40) &56.93(0.38)& -& -\\
german&69.99(0.27) &71.46(0.34)& -& -\\
glass& 42.97(2.66)&34.11(1.66)& -& -\\
heart-statlog&83.00(0.62)& 78.22(0.83)& -& -\\
iono& 93.28(0.68)&89.09(0.34)& -& -\\
sonar&86.11(1.37)& 68.50(2.13)& -& -\\
tae& 39.61(3.98)	&42.84(2.55)& -& -\\
vehicle& 45.39(0.61)&35.78(0.68)& -& -\\
wine& 95.11(0.65)&78.92(1.48)& -& -\\
wisc& 96.68(0.21)	&95.59(0.29)& -& -\\
zoo& 85.05(2.37)&88.30(5.09)& -& 86.90(2.45)\\
metabric&65.17(0.25) &51.81(0.29)& 43.18(0.66)& 67.14(0.25)\\
gse&92.86(0.59) &75.93(0.48)& 79.59(1.31)& 93.86(0.63)\\
3sources&71.79(1.33) &33.84(1.31)& -& 64.75(2.03)\\
gesture & 54.83(0.24) & 54.33(0.39) & - & - \\
   \bottomrule
   \end{tabular}
\label{table:additional}
\end{table}

\begin{table}[hpt]
	\centering
	\caption{Running time on the real-world datasets}
	\label{table:run_time1}
		\small	
 \begin{tabular}{lccccccc}
	\toprule
	Dataset&PCA&PPCA&GPLVM&KPCA(lin)&KPCA(gau)&KPCA(pol)&ICA\\
		\midrule
balance-scale & 0.0003 & 0.0985  & 2.4820  & 0.4598 & 0.4574 & 0.4187 & 0.0021 \\
german        & 0.0009 & 0.0576  & 27.0593 & 1.0942 & 0.7482 & 0.9136 & 0.1059 \\
glass         & 0.0003 & 0.0055  & 0.6492  & 0.0133 & 0.0366 & 0.0406 & 0.0032 \\
heart-statlog & 0.0005 & 0.0111  & 0.7930  & 0.0612 & 0.0552 & 0.0587 & 0.0034 \\
iono          & 0.0150 & 0.0956  & 6.0315  & 0.0122 & 0.0886 & 0.0862 & 0.0146 \\
sonar         & 0.0014 & 0.2056  & 1.4391  & 0.0347 & 0.0261 & 0.0411 & 0.0090 \\
tae           & 0.0003 & 0.0056  & 0.5483  & 0.0075 & 0.0211 & 0.0149 & 0.0071 \\
vehicle       & 0.0057 & 0.1332  & 4.0359  & 0.0851 & 0.7615 & 0.5744 & 0.0202 \\
wine          & 0.0003 & 0.0090  & 0.7537  & 0.0117 & 0.0348 & 0.0289 & 0.0052 \\
wisc          & 0.0027 & 0.0340  & 3.1442  & 0.0521 & 0.3989 & 0.4247 & 0.0331 \\
zoo           & 0.0004 & 0.0244  & 0.7022  & 0.0014 & 0.0125 & 0.0094 & 0.0080 \\
metabric      & 0.2006 & 65.8584 & 73.3401 & 0.6652 & 0.6625 & 6.3860 & 0.5514 \\
gse           & 0.0180 & 7.5660  & 3.6690  & 0.0661 & 0.0133 & 0.0527 & 0.0451 \\
3sources      & 0.0233 & 4.3406  & 1.3790  & 0.1291 & 0.0273 & 0.1052 & 0.1685\\
gesture  & 0.0013 & 0.0693  & 86.3266  & 10.2738 & 165.7107 & 106.9476 & 0.0372 \\
eeg\_eye & 0.0042 & 0.1493  & 208.5825 & 8.8485  & 215.5432 & 97.3119  & 0.0128 \\
mfeat    & 0.0674 & 23.6447 & 35.0592  & 11.0996 & 146.6688 & 66.5211  & 0.1396\\
		\bottomrule
	\end{tabular}
    \begin{tablenotes}
	\small
	\item  The running time is measured in seconds.
\end{tablenotes}
\label{table:rt1}
\end{table}

\begin{table}[hpt]
	\centering
	\caption{Running time on the real-world datasets (continue)}
	\small	
	\begin{tabular}{lccccccc}
		\toprule
		Dataset&GPCA(LL)&GPCA(LS)&	GPCA(SL)&GPCA(SS)&MN-PCA&MN-PCA-w2\\
		\midrule
balance-scale & - & 0.0291 & - & - & 0.7992   & 117.3487  \\
german        & - & 0.1457 & - & - & 1.8368   & 179.1464  \\
glass         & - & - & - & - & 0.2408   & 112.4802  \\
heart-statlog & - & 0.0047 & - & - & 0.1094   & 127.4794  \\
iono          & - & - & - & - & 2.2436   & 147.4035  \\
sonar         & - & 0.0328 & - & - & 0.1614   & 128.4263  \\
tae           & - & - & - & - & 0.0561   & 106.6672  \\
vehicle       & - & 0.0297 & - & - & 150.3426 & 142.4144  \\
wine          & - & 0.0054 & - & - & 0.0899   & 106.0842  \\
wisc          & - & - & - & - & 2.7413   & 138.6479  \\
zoo           & 0.0187 & - & 0.0039 & 0.0042 & 0.0357   & 100.3534  \\
metabric      & 1.9904 & 1.2931 & 1.2819 & 1.1250 & 111.1968 & 2384.8400 \\
gse           & 0.0442 & 0.0622 & 0.0370 & 0.0287 & 18.3070  & 251.8776  \\
3sources      & 0.2688 & - & 0.2288 & 0.1881 & 1.0401   & 363.4707 \\
geture   & - & 2.4268 & - & - & 343.4577  & 470.5875  \\
eeg-eye &- & - & - & - & 1778.3257 & 1261.2412 \\
mfeat    & - & - & - & - & 141.6516  & 927.1965\\
		\bottomrule
	\end{tabular}
\label{table:rt2}
\end{table}
\par
Table \ref{table:rt1} and \ref{table:rt2} report the running time of the algorithms on real-world datasets.
PCA and ICA are very efficient.
The precision matrix of GPCA is predefined.
Hence, the computational cost of GPCA is similar to PCA.
The probabilistic algorithms, including PPCA, GPLVM, MN-PCA and MN-PCA-w2, are relatively slow.
The proposed algorithms are slower because the estimation of the precision matrices is computationally expensive.
The time consumption of MN-PCA is heavily dependent on the non-zero entries in the precision matrices.
On the other hand, MN-PCA-w2 costs  more time because it involves SVD at each iteration and does not exploit the sparsity of the precision matrices.

\subsection{METABRIC Dataset}
We compare the estimated precision matrix in the sample space of QUIC and MN-PCA here.
To facilitate a fair comparison, we choose the $L_1$-norm parameter for QUIC such that the estimated precision matrix has the same number of edges approximately as that of MN-PCA.

\par
\begin{figure}[!htp]
    \centering
    \includegraphics[width=.75\textwidth]{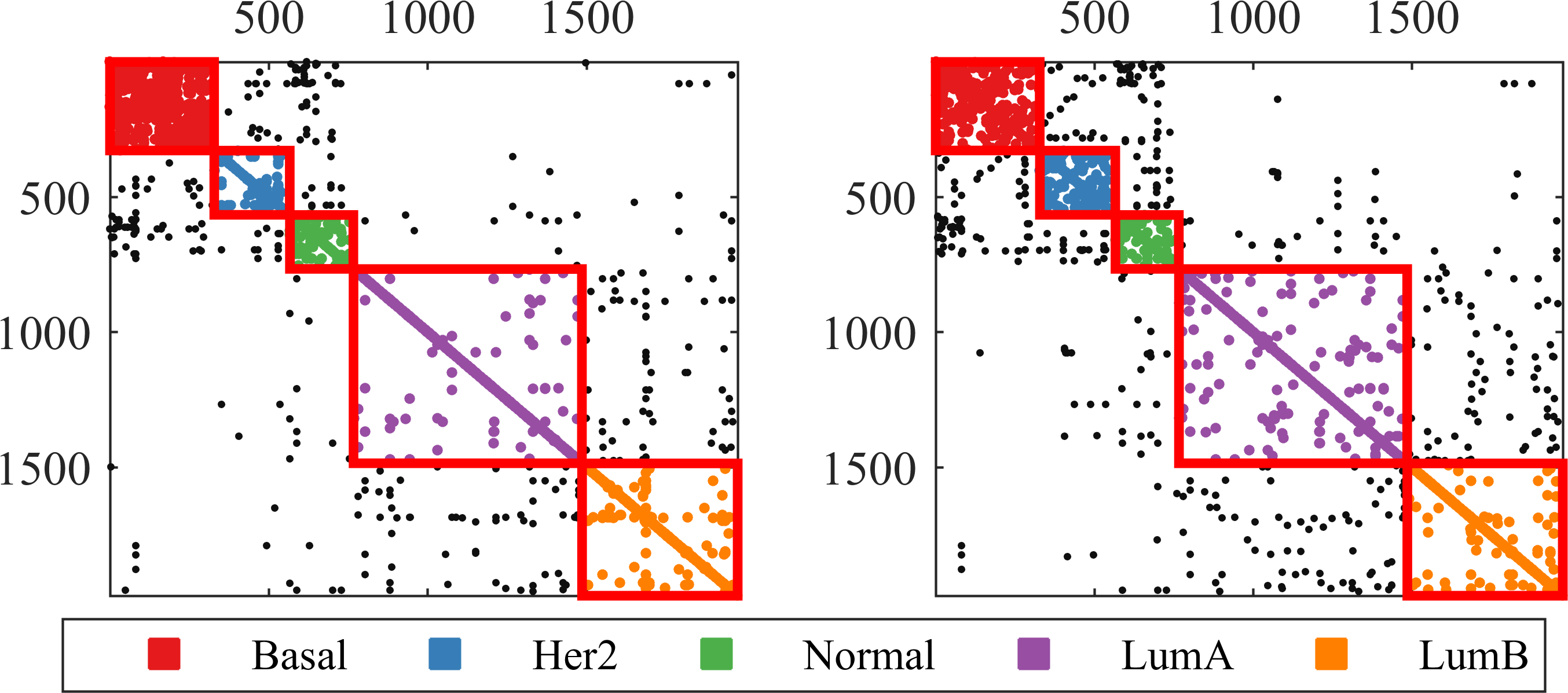}
    \caption{The estimated precision matrices of QUIC and MN-PCA in the sample space. They are reordered based on subtypes with $k = 5$. Only the top 1000 interactions are shown.}
    \label{fig:metabric_iA}
\end{figure}

 As shown in Fig. \ref{fig:metabric_iA}, the estimated precision matrices show similar patterns.

\subsection{3sources Dataset}
\par Here we compare the results of QUIC and MN-PCA on the 3source data. We choose the top 100 interactions in the estimated precision matrices in the feature (term) space by the absolute value.

\par Table \ref{table:3sources}  reports the selected terms of MN-PCA and QUIC.
$\text{MN-PCA} \cap \text{QUIC}$ indicates the term pairs that appear both in the top 20 of MN-PCA and QUIC. $\text{QUIC}\setminus \text{MN-PCA}$ indicates term pairs that appear in the top 20 of QUIC, but not in MN-PCA. Similarly, $ \text{MN-PCA}\setminus \text{QUIC}$ indicates term pairs that are in MN-PCA, but not in QUIC. We can see that all those term pairs, such as ``minister/primer", ``study/research", are semantic relative and thus they tend to appear in the same article. Besides, the top 20 interactions of QUIC and MN-PCA have considerable overlap, i.e., 11 out of 20. But they still show some differences.
\begin{table}[!ht]\caption{Estimated terms pairs on 3sources Data}\label{table:3sources}
    \setlength\extrarowheight{2pt} 
    \centering
    \begin{tabularx}{\columnwidth}{|X|X|}
        \hline
        \multicolumn{2}{|c|}{$ \text{MN-PCA}\cap \text{QUIC}$}\\
        \hline
        \multicolumn{2}{|p{.95\columnwidth}|}{minister/primer \ Gordon/prime \ Brown/prime \ Brown/Gordon \ tori/conserve \quad\quad \mbox{economy/economic} \ execut/chief \ university/research \ study/research \ premier/league \ journal/study
        }\\
        \hline
        $\text{QUIC}\setminus \text{MN-PCA}$  &  $ \text{MN-PCA}\setminus \text{QUIC}$ \\
        \hline
        party/conserve \ free/user \ internet/user \ site/user \ cup/football \ qualify/cup \ board/chairman \quad site/social \quad study/Dr   & G20/leader \ expense/MP \ \mbox{victory/win} \ club/football \ liverpool/league \ book/ref \mbox{referee/penalty}   financial/bank  \ cross/striker \\
        \hline
    \end{tabularx}
\end{table}

\bibliographystyle{IEEEtran}
\bibliography{IEEEabrv,reference}